\documentclass[letterpaper]{article}
\usepackage{aaai}
\usepackage{times}
\usepackage{helvet}
\usepackage{courier}
\usepackage{amssymb}
\usepackage{latexsym}

\usepackage{url}
\usepackage{xcolor}

\usepackage{natbib}
\usepackage{graphicx}
\usepackage{amsmath, amsfonts, dsfont}
\usepackage[bbgreekl]{mathbbol}
\usepackage{amsfonts}
\usepackage{enumitem}
\usepackage{footnote}
\usepackage{float}
\usepackage{hyperref}
\usepackage{pgfplots}
\pgfplotsset{compat=1.7}
\usepackage{times}
\usepackage{multicol}

\usepackage{array}
\usepackage{booktabs}
\usepackage{media9}
\usepackage{hyperref}

\usepackage{tikz}
\usepackage{forest}
\forestset{
  L1/.style={draw=black,},
  L2/.style={,edge={,line width=0.8pt}},
}

\usepackage{verbatim}
\usetikzlibrary{arrows,shapes}
\tikzstyle{format} = [draw, thin, fill=blue!20]
\tikzstyle{medium} = [ellipse, draw, thin, fill=green!20, minimum height=2.5em]
\usetikzlibrary{positioning,chains}
\usepackage{hyperref}

\usepackage[ruled,vlined]{algorithm2e}

\begin{document}
\title{A Novel Interaction-based Methodology Towards Explainable AI with Better Understanding of Pneumonia Chest X-ray Images}
\author{Shaw-Hwa Lo\\
Columbia University\\
shl5@columbia.edu\\
\And
Yiqiao Yin\\
Columbia University\\
yy2502@columbia.edu\\ }

\maketitle
\begin{abstract}
\begin{quote}
In the field of eXplainable AI (XAI), robust ``blackbox'' algorithms such as Convolutional Neural Networks (CNNs) are known for making high prediction performance. However, the ability to explain and interpret these algorithms still require innovation in the understanding of influential and, more importantly, explainable features that directly or indirectly impact the performance of predictivity. A number of methods existing in literature focus on visualization techniques but the concepts of explainability and interpretability still require rigorous definition. In view of the above needs, this paper proposes an interaction-based methodology -- Influence Score (I-score) -- to screen out the noisy and non-informative variables in the images hence it nourishes an environment with explainable and interpretable features that are directly associated to feature predictivity. We apply the proposed method on a real world application in Pneumonia Chest X-ray Image data set and produced state-of-the-art results. We demonstrate how to apply the proposed approach for more general big data problems by improving the explainability and interpretability without sacrificing the prediction performance. The contribution of this paper opens a novel angle that moves the community closer to the future pipelines of XAI problems.
\end{quote}
\end{abstract}

\section{Introduction}
\begin{figure}
    \centering
    \caption{This diagram is a recreation DARPA document (DARPA-BAA-16-53) \citep{DARPA2016} \citep{Rudin2019}. The diagram presents the relationship between learning performance (usually measured by prediction performance) and effectiveness of explanations (also known as explainability). The proposed method in our work aims to take any deep learning method and provide explainability without sacrificing prediction performance. In the diagram, this is marked as moving a dot (where a dot can be a machine learning algorithm) towards the upper right corner of the relationship plot.}
    \includegraphics[width=0.4\textwidth]{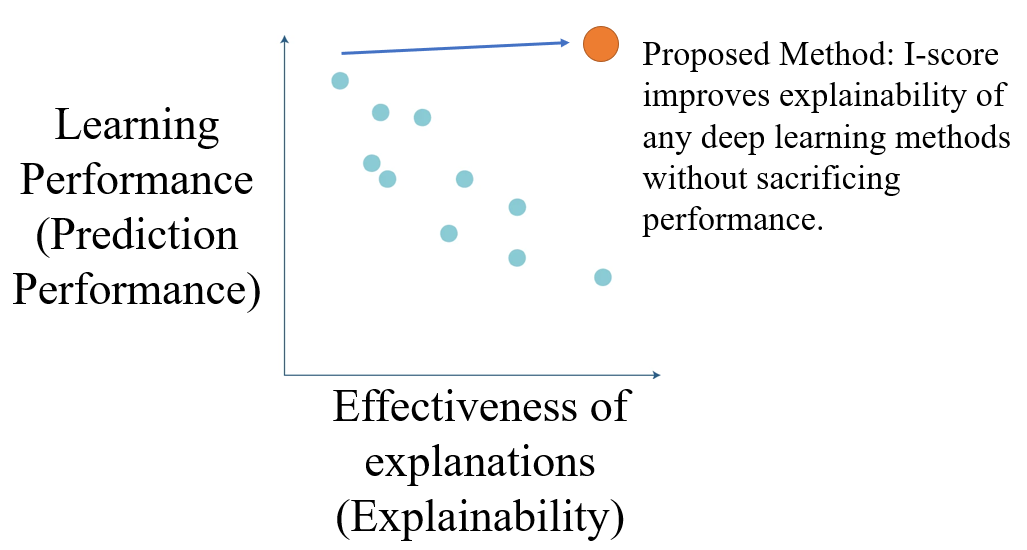}
    \label{fig:accuracy-explanation-tradeoff}
\end{figure}

Many successful achievements in machine learning and deep learning has accelerated real-world implementation of Artificial Intelligence (AI). This issue has been greatly acknowledged by the Department of Defense (DoD) \citep{DARPA2016}. In regarding to this need, \cite{DARPA2016} initiated the eXplainable Artificial Intelligence (XAI) challenge and brought this new interest to the surface. The guidance of XAI is to develop suites of explainable machine learning and deep learning methodologies to produce interpretable models that humans can understand, appropriately trust, and effectively manage the emerging generation of AI systems \citep{DARPA2016}. In addressing the concepts of interpretability and explainability, these scholars and researchers have made attempts towards discussing a trade-off between learning performance (usually measured by prediction performance) and effectiveness of explanations (also known as explainability) is presented in Figure \ref{fig:accuracy-explanation-tradeoff} \citep{linardatos2021} \citep{Lipton2018}. This trade-off often occurs in any supervised machine learning problems that aim to use explanatory variable to predict response variable (or outcome variable)\footnote{We use the terms response variable and outcome variable interchangeably throughout the article.} which happens between learning performance (also known as prediction performance) and effectiveness of explanations (also known as explainability). As illustrated in Figure \ref{fig:accuracy-explanation-tradeoff}, the issue is that a learning algorithm such as linear regression modeling has a clear algorithmic structure and an explicitly written mathematical expression so that it can be understood with high effectiveness of explanations with yet relatively lower prediction performance. This means that an algorithm such as linear regression can be positioned in the bottom right corner of the scale in this figure (in consensus, linear regression is regarded as explainable learning algorithm). On the other hand, a learning algorithm such as a deep Convolutional Neural Network (CNN) with hundreds of millions of parameters would have much better prediction performance, yet it is extremely challenging to explicitly state the mathematical formulation of the architecture. This means that a deep learning algorithm such as an ultra deep CNN with hundreds of millions of parameters would be positioned on the top left corner of the scale in the figure (which is generally considered inexplainable in consensus). \textbf{In the field of transfer learning, it is a common practice to adopt a previously trained CNN model on a new data set. For example, one can adopt the VGG16 model and weights learned from ImageNet on a new data: Chest X-ray Images. The filters learned from ImageNet data may or may not be helpful on Chest X-ray. Due to large amount of filters used in VGG16, we can hope that some filters can capture important information on Chest X-ray scans. However, we will never truly know what features are important if we do not impose any feature assessment condition. This renders the adoption of a pretrained CNN model inexplicable.} This calls for the need of a novel feature assessment and feature selection technique to shrink the dimension of the number of parameters while maintaining prediction performance. Hence, this paper focuses on feature and variable selection assessment to build explainability including trustworthy, fair, robust, and high performing models for real-world applications. A fruitful consequence of this delivery is to build learning algorithms with state-of-the-art performance while maintaining small number of features and parameters, an algorithm that can be positioned on the top right corner of the scale in Figure \ref{fig:accuracy-explanation-tradeoff}.

\section{Definition of Feature Explainability}
A popular description of interpretability is by \cite{DoshiVelez2017TowardsAR} which defined XAI as the ability to explain or to present in understandable terms to a human. Another popular version states interpretability as the degree to which a human can understand the cause of a decision \citep{MILLER20191}. Though intuitive, these definitions lack mathematical formality and rigorousness \citep{Adadi2018}. Moreover, it is yet unclear why variables provide us the good prediction performance and, more importantly, how to yield a relatively unbiased estimate of a parameter that is not sensitive to noisy variables and is related to the parameter of interest. Insofar, there has not been any thorough notions defined about the explainability of features or explanatory variables. Due to this issue, it is imminent for the literature to develop rigorous definition and concept for explainable and interpretable feature assessment techniques. We regard the outcome measure of an explainable feature assessment and selection methodology the explainability of explanatory variables. With these techniques identified, we can appropriately address the explainability of a variable set using the explainable and interpretable feature assessment methodologies.

To shed light to these questions, we define the following three necessary conditions ($\mathcal{C}1$, $\mathcal{C}2$, and $\mathcal{C}3$) for any feature selection methodology to be explainable and interpretable. \textbf{In other words, a variable and feature selection method can only be considered explainable and interpretable if all three conditions ($\mathcal{C}1$, $\mathcal{C}2$, and $\mathcal{C}3$ defined below) are satisfied.} Specifically, we regard the final assessment quantity of the importance evaluated for a set of features or variables to be the final score measured by feature assessment and selection method. \textbf{More importantly, we define this importance score of a variable set from using only explainable feature assessment and selection methods to be the explainability of a combination of variables. There are three conditions defined below and we name these conditions $\mathcal{C}1$, $\mathcal{C}2$, and $\mathcal{C}3$.}
\begin{itemize}
    \item $\mathcal{C}1$. The first condition states that the feature selection methodology do not require the knowledge of the underlying model of how explanatory variables affects outcome variable. This means the following. Suppose a prediction task has explanatory variables $X$ and outcome variable $Y$. If we fit a model $f(\cdot)$ and we use $f(X)$ to explain $Y$, the procedure requires us to explicitly understand the internal structure of $f(\cdot)$. This is extremely challenging to establish a well-written mathematical formulation when the fitted model is an ultra-deep CNN with hundreds of millions of parameters (sometimes it is combination of many deep CNNs). Worse yet, any attempted explanations would have mistakes of this model $f(\cdot)$ carried over. To avoid this bewilderment scientists and statisticians feel in explaining outcome variable, a feature selection methodology is required to understand how explanatory variables $X$ impact the outcome variable $Y$ without relying on the procedure of searching for a fitted model $f(\cdot)$. Hence, the first condition, $\mathcal{C}1$, requires the feature selection methodology to not depend on the model fitting procedure (to avoid the procedure of searching for $f(\cdot)$).
    \item $\mathcal{C}2$. An explainable and interpretable feature selection method must clearly state to what degree a combination of explanatory variables influence the response variable. Moreover, it is beneficial if a statistician can directly compute a score for a set of variables in order to make reasonable comparisons. This means any additional influential variables should raise this score while any injection of noisy and non-informative variables should decrease the value of this score. Hence, this condition, $\mathcal{C}2$, allows statisticians to pursue feature assessment and feature selection in a quantifiable and rigorous manner. Since we consider the explainability to be the final score and assessment of a variable set using only explainable feature assessment methodology, this second condition, $\mathcal{C}2$, asserts that explainability of a combination of variables to be exactly the amount of assessment that explainable feature assessment and selection methodology evaluates and it is a concept states how important explanatory variables are at influencing outcome variable.
    \item $\mathcal{C}3$. In order for a feature assessment and selection technique to be interpretable and explainable, it must directly associate with the predictivity of the explanatory variables (for definition of predictivity, please see \cite{lochernoffzhenglo2015} \cite{lochernoffzhenglo2016}). This is because in deep learning era the variables are commonly explained in two ways. The first is to rely on the weights (also known as the parameters) found by backpropagation. For example, consider input variables (or explanatory variables) to be $X_1$ and $X_2$. A very simple neural network architecture can be constructed using weights $w_1$ and $w_2$. We can simply define estimated outcome variable $Y$ to be $\hat{Y} := \text{sigmoid}(\sum_{j=1}^2 w_jX_j) = (1 + \exp(- \sum_{j=1}^2 w_j X_j))^{-1}$. Though with little intuition, $w_j$ is commonly used to illustrate how much $X_j$ affects $Y$. The second is through visualization after model fitting procedure. For example, CAM (and its upgraded versions) \citep{Zhou2016} can be used to generate highlights of images that are important for making predictions. The first approach of using weight parameters can be quite challenging and even an impossible task when the neural network architecture has hundreds of hidden layers and hundreds of millions of parameters. The second approach, however, requires the statistician to have the access of internal structure of the model. This would be difficult if an ultra-deep CNN is used with millions of parameters. In addition, the approaches above fail to associate with the predictivity of a particular variable set and they violate the first condition, $\mathcal{C}1$, as well.
\end{itemize}

Only with all three conditions satisfied, a feature assessment and selection technique would be considered explainable and interpretable in this article. We regard these three conditions to be required in order for a feature assessment and selection methodology to be considered explainable and interpretable. In addition, we consider the explainability of a variable set to be exactly the outcome score of an explainable feature assessment and selection method. Only with this appropriate method that satisfies the definition of explainable feature assessment technique can we say how the explanatory variables explains the outcome variable. Hence, we propose a research agenda of feature selection methodology that evaluates the explainability of explanatory variables. 

\begin{figure}
    \centering
    \caption{The graph presents two samples using CAM visualization technique before and after using proposed method. This marks the direction of explainability that this paper tackles. We will see more intriguing results in the application section.}
    \includegraphics[width=0.47\textwidth]{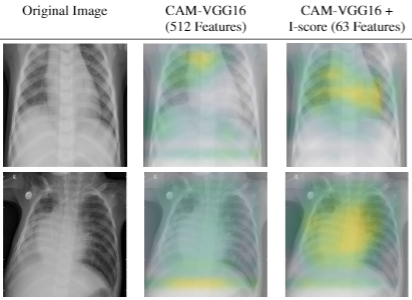}
    \label{fig:cam-VGG16-two-samples}
\end{figure}

With these questions in front of us, research agenda towards searching for a criterion to locate highly predictive variables is imminent. \cite{Bam2019} raised the question of absolute feature importance (exactly how important each feature is), but it is yet unexplored how to search for important features by directly looking at the given data set before fitting a model which fails to check the first condition, $\mathcal{C}1$. Amongst a variety of deep learning frameworks, the Convolutional Neural Networks (CNNs) \citep{LeCun89} have been widely adopted by many scholarly work including computer vision, object detection \citep{Girshick2014} \citep{Gordo2016}, image recognition \citep{He2006} \citep{Krizhevsky2012} \citep{Szegedy2015}, image retrieval \citep{Gong2014} \citep{Gordo2016}, and so on. Many famous network architectures that exist including VGG16 \citep{Simonyan2014}, VGG19 \citep{Simonyan2014}, DenseNet121 \citep{Huang2016}, DenseNet169 \citep{Huang2016}, DenseNet201 \citep{Huang2016}, ResNet \citep{He2016}, and Xception \citep{Chollet2016}. Despite their \citep{Chollet2016}, \citep{He2016}, \citep{Huang2016}, \citep{Krizhevsky2012}, \citep{Simonyan2014} remarkable prediction performance, some have proposed to avoid the use of fully-connected layers to minimize the number of parameters while maintaining high performance \citep{Zhou2016}. In the work of  \cite{Zhou2016}, experiments show advantages of using global average pooling layer to retain localization ability until the final layer. However, these doctrines require the access of internal structure of a CNN and, hence, fail to meet the first condition, $\mathcal{C}1$. \cite{Zhou2016} provided explanation of the important characteristics of regions in the image emphasized by CNNs for determining the classification of the entire image. However, weights of pooling layer, determined by backpropgation, may also take background information into consideration when the algorithm is making predictions. This may produce inconsistency with human beliefs since humans generally tend to focus on the foreground characteristics which do not meet the second condition, $\mathcal{C}2$. For example, the investigation in this paper used Class Activation Map (CAM) on the X-ray images of Pneumonia diseased patients. Human experts generally study the lung area and explores how disease manifests. However, an AI-trained model may use background area (such as liver, stomach, and intestines) instead of the foreground (features of lungs) to make predictions (see Figure \ref{fig:cam-VGG16-two-samples}).

The above methods all focus on variable selection and feature importance ranking. However, no method has been able to provide sound approach to meet all three conditions: $\mathcal{C}1$, $\mathcal{C}2$, and $\mathcal{C}3$ described above. Specifically, it is not yet fully discovered how to select variables without assuming any model formation from many noisy variables and, moreover, how to detect, out of many potential explanatory features, the most influential features that directly have impact on response variable $Y$. 

\cite{chernoffetal2009} presents a general intensive approach, based on a method pioneered by \cite{lozheng2002} for detecting which, out of many potential explanatory variables, have an influence given a specific subset of explanatory variables on a dependent variable $Y$. This paper presents an interaction-based feature selection methodology incorporating the notion of influence score, I-score, as a major technique to detect the higher-order interactions in complex and large-scale data set. Our work also investigates the potential usage of I-score to explain and visualize the deep learning framework. It is not surprising that these new tools (I-score and Backward Dropping Algorithm) are able to provide vast explanation directly associated to response variable and interpretation to visualize any given CNNs and other deep learning frameworks.

\section{Proposed Method}
The proposed methodology comes with three stages. First, we investigate variables to identify those with high potential to form influential modules. Secondly, we generate highly influential variable modules from variables selected in the first stage, where variables of the same module interact with each other to produce a strong effect on $Y$. Last, we combine the variable modules to carry out prediction process.

From prior simulation experience \citep{lochernoffzhenglo2015} \citep{lochernoffzhenglo2016}, it is demonstrated that the two basic tools, I-score and Backward Dropping Algorithm, can extract influential variables from data set in regards of modules and interaction effect. However, questions remain on how to determine the input to Backward Dropping Algorithm and how to use the output from Backward Dropping Algorithm results to construct prediction estimates. Unless one can appropriately utilize input to and output from Backward Dropping Algorithm, the strength of Backward Dropping Algorithm cannot be fully excavated. In this sense, the innovation of the proposed method manifests itself in three ways. First, if one directly applies Backward Dropping Algorithm on high-dimensional data sets, one may still miss some useful information. We propose two-stage feature selection procedure: interaction-based variable screening and variable module generation via Backward Dropping Algorithm. Since the impurity of features is largely enhanced by the interaction-based variable selection algorithm in the first stage, we are able to construct variable modules that are higher order interactions with lots of information in the second stage. These variable modules will support as building blocks for us to form classification rules. This school of thoughts produce results significantly better than directly application of Backward Dropping Algorithm.

The statistics, I-score, is defined using discrete variables. If some explanatory variables are continuous, we first convert them into discrete ones for feature selection purpose (see \S2.3.3 for Artificial Example III for detailed discussion of discretization). After we have selected the important variables, we can use the original values to estimate their effects. We can rely on the influence score when we convert continuous variables into discrete variables. For example, if a random variable is drawn from normal distribution, then one optimal cutoff is to use the one that has the largest marginal I-score. There is a trade-off induced from this process: the information loss due to discretizing variables from continuous to discrete forms versus the information gain from robust detection of interactions by discretization. Wang, Lo, Zheng, and Hu (2002) \citep{wanglozhenghu2012} demonstrated that the gain from robust detection of interactions is much more than enough to offset possible information loss due to discretization. Wang, Lo, Zheng, and Hu (2002) \citep{wanglozhenghu2012} used the two-mean clustering algorithm to turn the gene expression level into a variable of two categories, high and low. As an additional piece of evidence supporting the proposed pre-processing step, the authors have tried more than two categories; e.g. three categories of high, medium and low. The empirical results show that the more categories used the worse classification error rates.

\subsection{Influence Score (I-score)}
The Influence Score (I-score) is a statistic derived from the partition retention method \citep{chernoffetal2009}. Consider a set of $n$ observations of an outcome variable (or response variable) $Y$ and a large number $S$ of explanatory variables, $X_1$, $X_2$, ... $X_S$. Randomly select a small group, $m$, of the explanatory variables $X$'s. We can denote this subset of variables to be $X\{X_k, k = 1, ..., m\}$. We suppose $X_k$ takes values of only 1 and 0 (though the variables are binary in this discussion, it can be generalized into continuous variables, see \S3.3.3 Artificial Example III). Hence, there are $2^m$ possible partitions for $X$'s. The $n$ observations are partitioned into $2^m$ cells according to the values of the $m$ explanatory variables. We refer to this partition as $\Pi_X$. The proposed I-score (denoted by $I_{\Pi_X}$) is defined in the following
\begin{equation}
    I_{\Pi_X} = \frac{1}{n s_n^2} \sum_{j=1}^{2^m} n_j^2 (\bar{Y}_j - \bar{Y})^2
\end{equation}
while $s_n^2 = \frac{1}{n} \sum_{i=1}^n (Y_i - \bar{Y})^2$. We notice that the I-score is designed to capture the discrepancy between the conditional means of $Y$ on $\{X_1, X_2, ..., X_m\}$ and the mean of $Y$. 

The statistics $I$ is the summation of squared deviations of frequency of $Y$ from what is expected under the null hypothesis. There are two properties associated with the statistics $I$. First, the measure $I$ is non-parametric which requires no need to specify a model for the joint effect of $\{X_{b_1}, ..., X_{b_k}\}$ on $Y$. This measure $I$ is created to describe the discrepancy between the conditional means of $Y$ on $\{X_{b_1}, ..., X_{b_k}\}$ disregard the form of conditional distribution. With each variable to be dichotomous, the variable set $\{X_{b_1}, ..., X_{b_k}\}$ form a well-defined partition, $\mathcal{P}$ \citep{chernoffetal2009}. Secondly, under the null hypothesis that the subset has no influence on $Y$, the expectation of $I$ remains non-increasing when dropping variables from the subset. The second property makes $I$ fundamentally different from the Pearson's $\chi^2$ statistic whose expectation is dependent on the degrees of freedom and hence on the number of variables selected to define the partition. We can rewrite statistics $I$ in its general form when $Y$ is not necessarily discrete
\begin{equation}\label{eq:iscore}
I = \sum_{j \in \mathcal{P}} n_j^2 (\bar{Y}_j - \bar{Y})^2,
\end{equation}
where $\bar{Y}_j$ is the average of $Y$-observations over the $j$th partition element (local average) and $\bar{Y}$ is the global average. Under the same null hypothesis, it is shown (Chernoff et al., 2009 \citep{chernoffetal2009}) that the normalized $I$, $I/n\sigma^2$ (where $\sigma^2$ is the variance of $Y$), is asymptotically distributed as a weighted sum of independent $\chi^2$ random variables of one degree of freedom each such that the total weight is less than one. It is precisely this property that serves the theoretical foundation for the following algorithm.

We further discuss in the Artificial Example II of the section \ref{supplement} the comparison between AUC values and I-score in a simulated environment.

\subsection{Backward Dropping Algorithm (BDA)}
The Backward Dropping Algorithm is a greedy algorithm to search for the optimal subsets of variables that maximizes the I-score through step-wise elimination of variables from an initial subset sampled in some way from the variable space. The steps of the algorithm are as follows.
\begin{algorithm}\label{alg:BDA}
\SetAlgoLined
\caption{Procedure of the Backward Dropping Algorithm (BDA)}
\emph{Training Set}: Consider a training set $\{(y_1, x_1), ..., (y_n, x_n)\}$ of $n$ observations, where $x_i = (x_{1i}, ..., x_{pi})\}$ is a $p$-dimensional vector of explanatory variables. The size $p$ can be very large. All explanatory variables are discrete.\;
\emph{Sampling from Variable Space}: Select an initial subset of $k$ explanatory variables $S_b = \{X_{b_1}, ..., X_{b_k}\}$, $b = 1, ..., B$\;
    \emph{Initialization}: Set $l = k$\;
    \While{While $l >= 1$}{
        \emph{Compute Standardized I-score}: calculate $I(S_b) = \frac{1}{n\sigma^2}\sum_{j \in \mathcal{P}_k} n_j^2 (\bar{Y}_j - \bar{Y})^2$. For the rest of the paper, we refer this formula as Influence Measure or Influence Score (I-score). \\
        \emph{Drop Variables}: Tentatively drop each variable in $S_b$ and recalculate the I-score with one variable less. Then drop the one that gives the highest I-score. Call this new subset $S_b'$ which has one variable less than $S_b$. \\
        l = $|S_b'|$ (update $l$ with length of current subset of variables)
    }
\end{algorithm}

We discuss in details in the Artificial Example I of the section \ref{supplement} and we illustrate the procedure of using the BDA in a simulated environment.

\section{Application}
This section presents the results of the experiments. 

\subsection{Background of the Pneumonia Disease}
Pneumonia has been playing a major component of the children morality rate across the globe. According to statistics from World Health Organization (WHO), an estimated of 2 million deaths are reported every year for children under age 5. In the United States, pneumonia accounts for over 500,000 visits to emergency departments and 50,000 deaths in 2015, keeping the ailment on the list of the top 10 causes of death in the country. Chest X-ray (CXR) analysis is the most commonly performed radiographic examination for diagnosing and differentiating the types of pneumonia \citep{Cherian2005}. While common, accurately diagnosing pneumonia is possible with modern day technology, it is a requirement to review chest radiograph (CXR) by highly trained specialists and confirmation through clinical history, vital signs and laboratory examinations. Pneumonia usually manifests as an area or areas of increased opacity on CXR \cite{Franquet2018}.  However, the diagnosis of pneumonia on CXR is complicated because of a number of other conditions in the lungs such as fluid overload (pulmonary edema), bleeding, volume loss (atelectasis or collapse), lung cancer, or post-radiation or surgical changes. Outside of the lungs, fluid in the pleural space (pleural effusion) also appears as increase opacity on CXR. \cite{Togacar2019} used a deep CNN with 60 million parameters. \cite{Ayan2019} tested an ultra deep convolutional neural network that has 138 million parameters. \cite{Rajaraman2018} attempted a variety of architecture that ranges in 0.8 - 40 million parameters. While some of these architecture produced good prediction performance, the power of explainability is lost in these deep convoluted structures. The proposed methods, however, do not rely on deep convoluted network architecture and adopt single-layer network with a single pass of propagation.

\begin{figure*}[!t]
\centering
\caption{The following graph shows randomly sampled images from pediatric CXRs collected in Pneumonia data set. Panel A presents 9 images from Class Diseased. These CXR show diseased pictures with bacterial affected pneumonia disease. Panel B presents 9 images from Class Normal. These images show clear lungs with no abnormal opacity.}
\begin{tabular}{cc}
\includegraphics[width=0.4\textwidth]{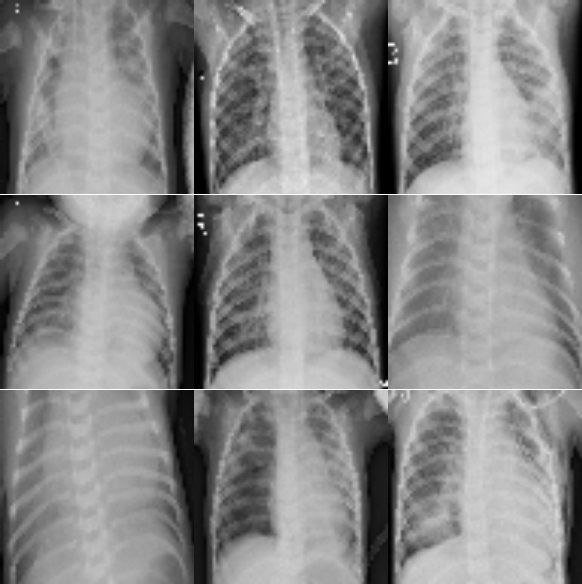} & 
\includegraphics[width=0.4\textwidth]{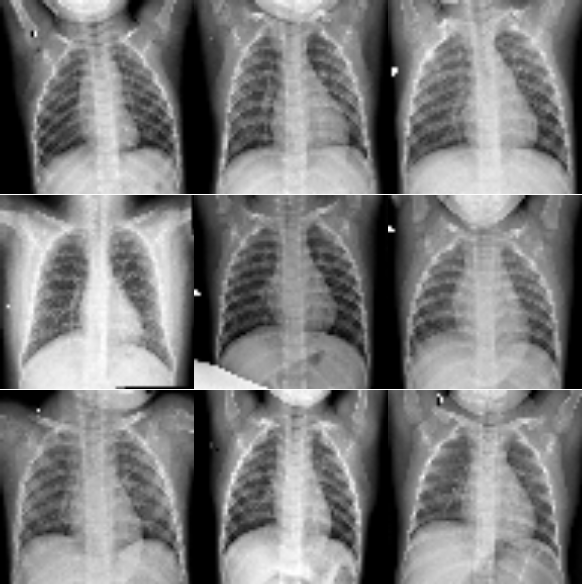} \\
Panel A & Panel B \\
\end{tabular}
\label{fig:pneumonia-data-sampled-images}
\end{figure*}

\subsection{Biological Interpretation of the Image Data}
An important notion is Opacity. Opacity refers to any area that preferentially attenuates the x-ray beam and therefore appears more opaque than the surrounding area. It is nonspecific term that does not indicate the size or pathological nature of the abnormality.\footnote{Felson's Principles of the Chest Roentgenology (4E), available from \url{https://www.amazon.com/Felsons-Principles-Roentgenology-Programmed-Goodman/dp/1455774839}.} From observation, lung opacity is not homogeneous and it does not have a clear center or clear boundaries. There is no universal methodology to properly segment opacity out of the entire picture. However, if one can segment the lungs properly and filter out the rest of the image, one could create a clean image of the lungs for neural network to process. 

\begin{figure}[!t]
\centering
\caption{This image shows the lung anatomy and use graphical diagram to illustrate how gas exchange cycles in healthy human lungs versus diseased human lungs.}
\includegraphics[width=0.47\textwidth]{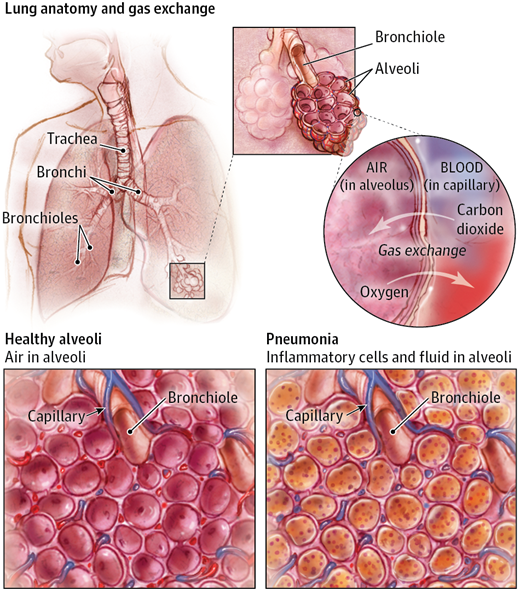}
\label{fig:pneumonia-data-lung-anatomy}
\end{figure}

There is a biological reason of why different healthy level of lungs exhibit different level of opacity. To illustrate this idea, a diagram of lung anatomy and gas exchange is posed in the following (see Figure \ref{fig:pneumonia-data-lung-anatomy}). The structure of the human lungs consists of Trachea, Bronchi, Bronchioles, and Alveoli. The most crucial activity is the cycled called Gas Exchange. Trachea acts like a main air pipe allowing air to pass through from outside of human body to inside human chest area. The Bronchi connects Trachea and are thinner pipes that allow air to move further into lung area. The end of Bronchi has many tiny airbags called Alveoli. Alveioli is the center location for gas exchange and it has a thin membrane to separate air and bloodstream. As human beings conduct day-to-day activities such as running, walking, or even sleeping, bloodstream is constantly filled with Carbon Dioxide that is generated from these activities which then need to be passed out of the human body. The pass from bloodstream into Alveoli is the first step. The reverse direction also has an activity for Oxygen to pass into the bloodstream so human beings can continue to conduct normal day-to-day behavior. The in-and-out cycle with Carbon Dioxide and Oxygen is called the Gas Exchange which is a normal microscopic activity occurs disregard whether human beings are conscious or not. For patients with diseased lungs, it is a natural reaction that the immune system is fighting against the germs and the inflammatory cells. This progress creates fluids inside Alveoli which generates grey area on image because not all CXR go through lungs. This creates the opaqueness in the images collected from CXR machine. 

Neuman et al. (2012) \citep{Neuman2012} proposed methods to evaluate radiography of children in a pediatric emergency department for suspicion of pneumonia. A team of six radiologists at two academic children's hospitals were formed to examine the image data of the chest x-ray pictures. How accurate are the annotated boxes? Neuman's work suggested there was only a moderate level of agreement between radiologists about the presence of the opacity.

\begin{figure}[!t]
\centering
\caption{This image shows the lung anatomy and several annotated boxes doctors use to indicate important area in the image for diagnosis. Neuman et al. (2012) \citep{Neuman2012} suggested such annotation can be helpful for Pneumonia diagnosis.}
\includegraphics[width=0.47\textwidth]{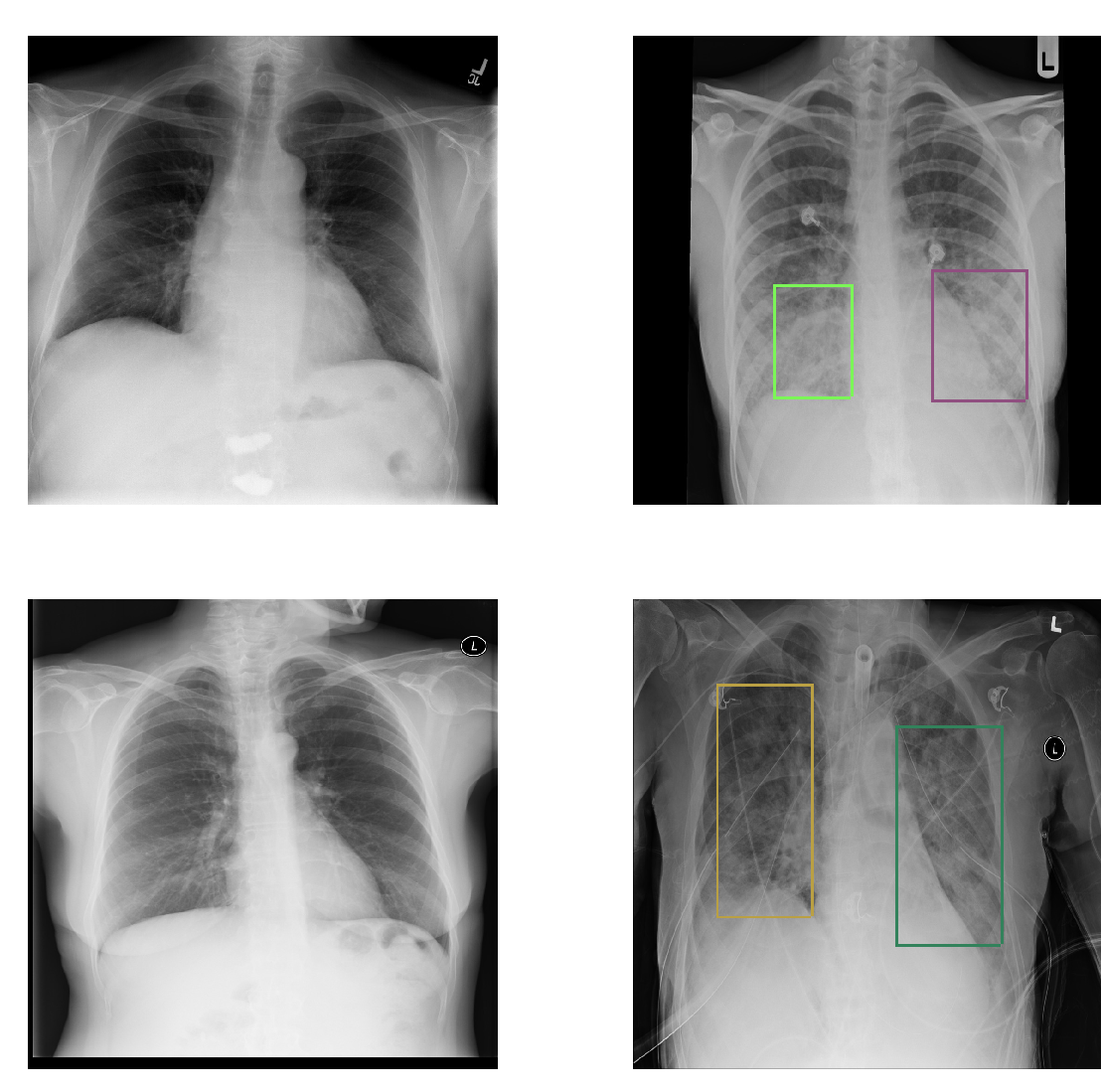}
\label{fig:pneumonia-data-lung-anatomy-modern}
\end{figure}

\subsection{Pneumonia Data Set}
The Pneumonia data set has images from different sizes (usually range from 300 to 400 pixels in width). For the experiment analyzing pixels, we reshape all images into 3-dimensional tensor that has size 224 by 224 by 3. In other words, each image has 150,528 pixels taking values from 0 to 255 before rescaling. There are 1,341 images with label normal class and 3,875 images with label pneumonia. In other words, this is a binary classification problem and we want to train a machine to learn the features of the images to be able to predict a probability that an image fall in class normal or pneumonia. We randomly select 300 images from normal cases and 300 images from diseased cases as test data. We then sample with replacement 3,000 images from normal cases and 3,000 images from diseased cases as training. This statistics can be summarized in the following table (see Table \ref{tab:pneumonia-data-train-test-split}).

\begin{table}[!t]
\centering
\caption{The following table presents the size of the training set, the validating set, and the test set for both approaches. The proposed approach is Interaction-based Feature Learning which proposes to work with convolutional features instead of the original pixels.}
\begin{tabular}{llccc}
\toprule
Proposed Approach & Normal & Pneumonia \\
Size: 224 by 224 by 3 & Obs. for & Obs. for \\
\hline
Training & 2,700 & 2,700 \\
Validating & 300 & 300 \\
Testing & 300 & 300 \\
\bottomrule
\end{tabular}
\label{tab:pneumonia-data-train-test-split}
\end{table}

\subsection{Transfer Learning Using VGG16}

To borrow the strength of another deep Convolutional Neural Network, transfer learning is a common scheme in adopting many pre-trained models in a new data set. However, due to significantly large amount of filters, it is much less optimal if the prediction performance lack of explainability. 

\begin{figure}
    \centering
    \caption{This figure represents the VGG16 Architecture \cite{Simonyan2014}.}
    \includegraphics[width=0.47\textwidth]{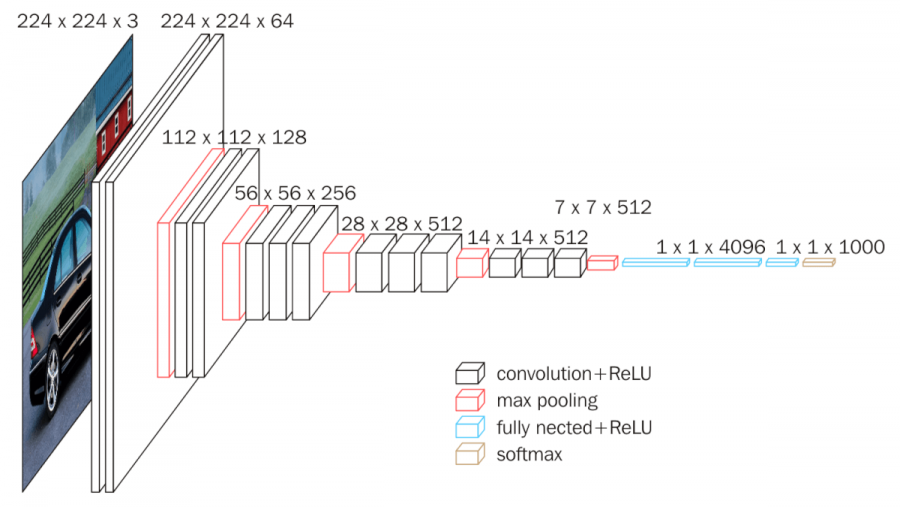}
    \label{fig:vgg16-architecture}
\end{figure}

\begin{figure}
    \centering
    \caption{This figure presents filters extracted from the architecture of VGG16 model. There are 64 filters in VGG16 model and they are presented above. Each filter has size 3 by 3 by 3. We code them into three colors by using RGB palette.}
    \includegraphics[width=0.47\textwidth]{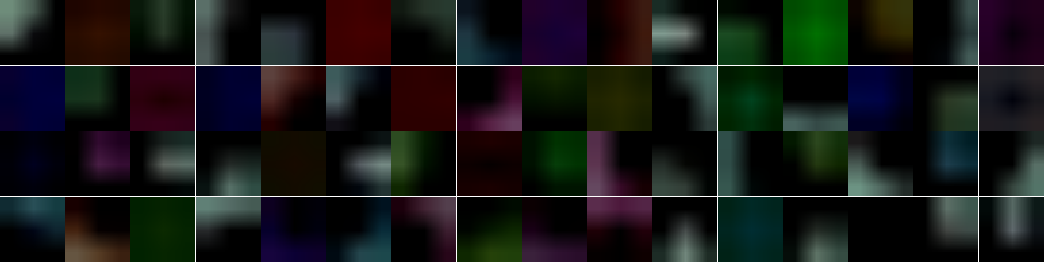}
    \label{fig:VGG16-filters}
\end{figure}

\begin{figure}
    \centering
    \caption{This figure represents the resulting images from each layer of the VGG16 architecture \cite{Simonyan2014}. Given a new data set (i.e. the Pneumonia Chest X-ray dataset), we can feed each image into the VGG16 architecture to create feature maps from using the parameters and filter kernels in the original VGG16 model.}
    \begin{tikzpicture}
        \begin{scope}[cm={1,0.7,0,1,(0,0)}]
            \node[transform shape, draw, black, ultra thick, inner sep=0.2mm] {\includegraphics[width=.17\textwidth]{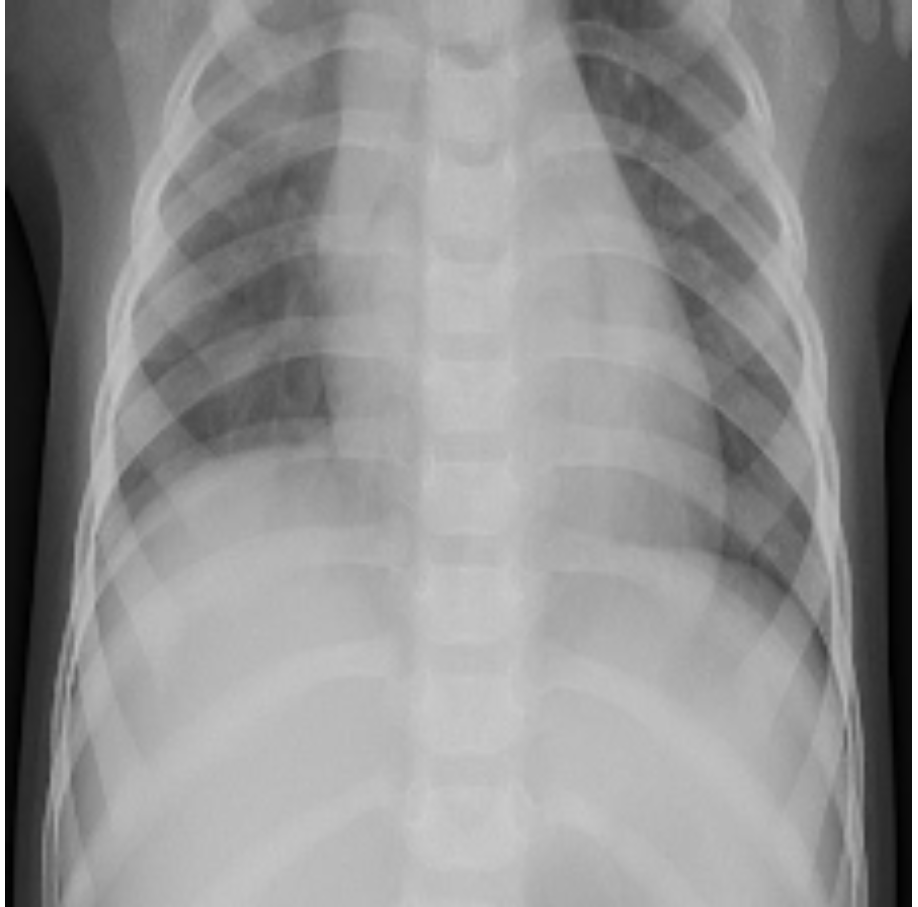}};
        \end{scope}
        
        \begin{scope}[cm={1,0.7,0,1,(1.5,0)}]
            \node[transform shape, draw, black, ultra thick, inner sep=0.2mm] {\includegraphics[width=.2\textwidth]{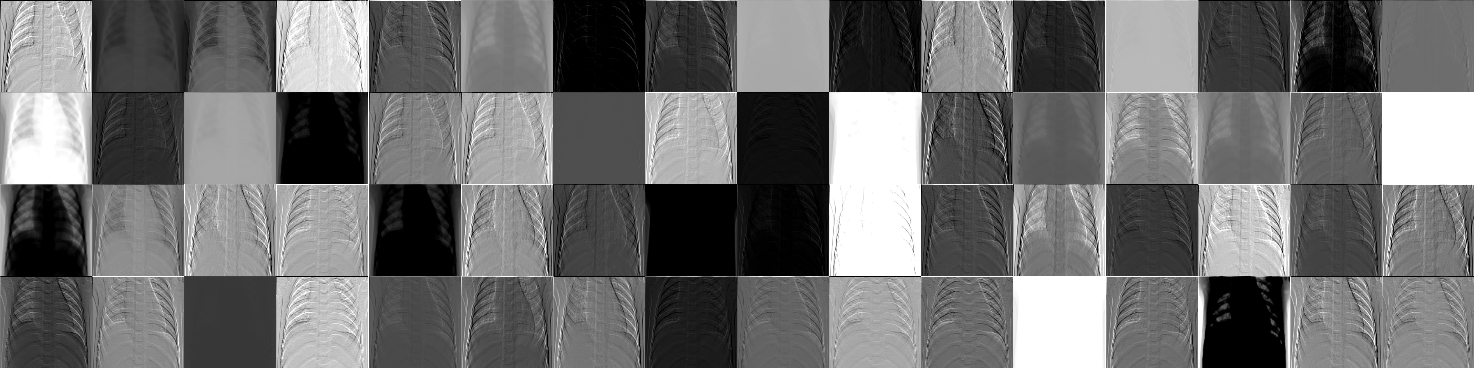}};
        \end{scope}
        \begin{scope}[cm={1,0.7,0,1,(1.6,0)}]
            \node[transform shape, draw, black, ultra thick, inner sep=0.2mm] {\includegraphics[width=.2\textwidth]{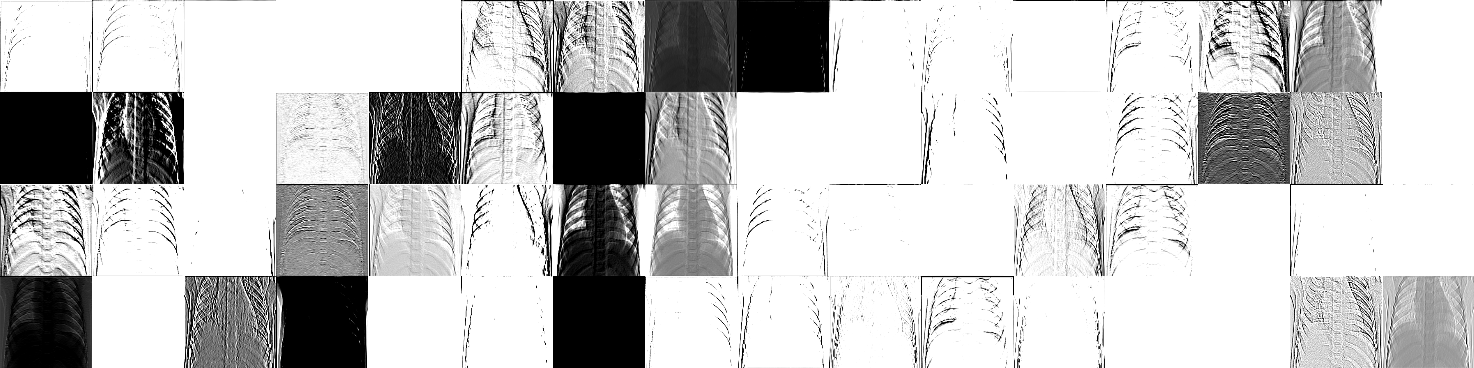}};
        \end{scope}
        
        \begin{scope}[cm={1,0.7,0,1,(2.4,0)}]
            \node[transform shape, draw, black, ultra thick, inner sep=0.2mm] {\includegraphics[width=.1\textwidth]{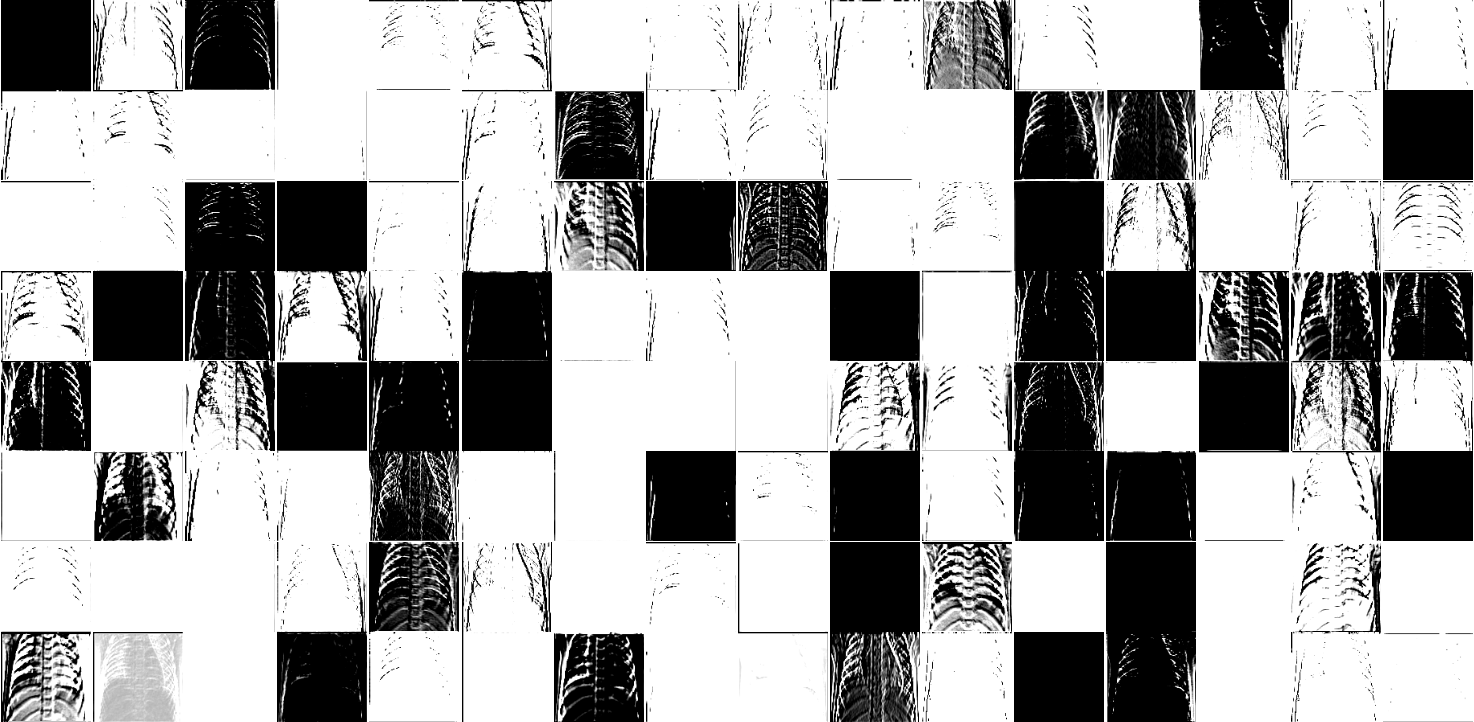}};
        \end{scope}
        \begin{scope}[cm={1,0.7,0,1,(2.5,0)}]
            \node[transform shape, draw, black, ultra thick, inner sep=0.2mm] {\includegraphics[width=.1\textwidth]{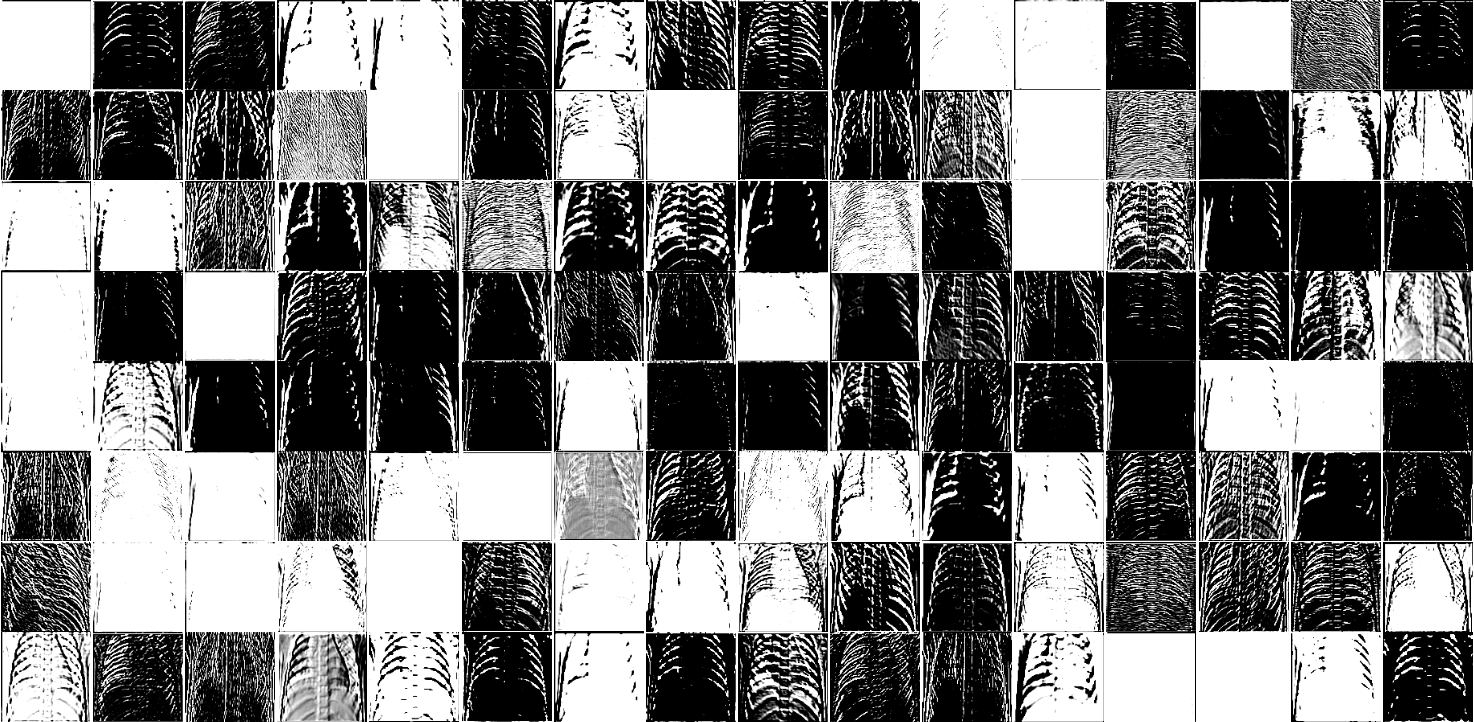}};
        \end{scope}
        
        \begin{scope}[cm={1,0.7,0,1,(3.3,0)}]
            \node[transform shape, draw, black, ultra thick, inner sep=0.2mm] {\includegraphics[width=.1\textwidth]{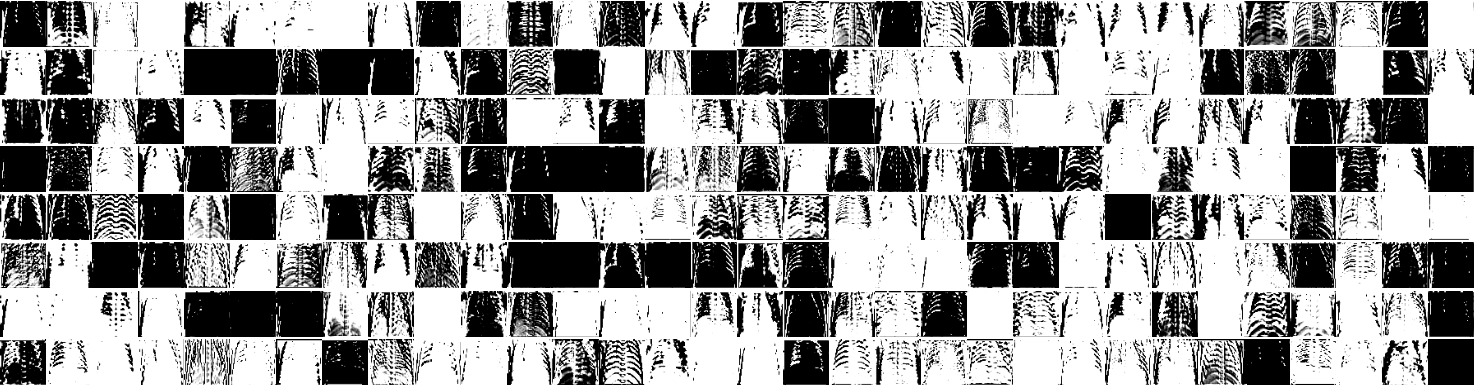}};
        \end{scope}
        \begin{scope}[cm={1,0.7,0,1,(3.5,0)}]
            \node[transform shape, draw, black, ultra thick, inner sep=0.2mm] {\includegraphics[width=.1\textwidth]{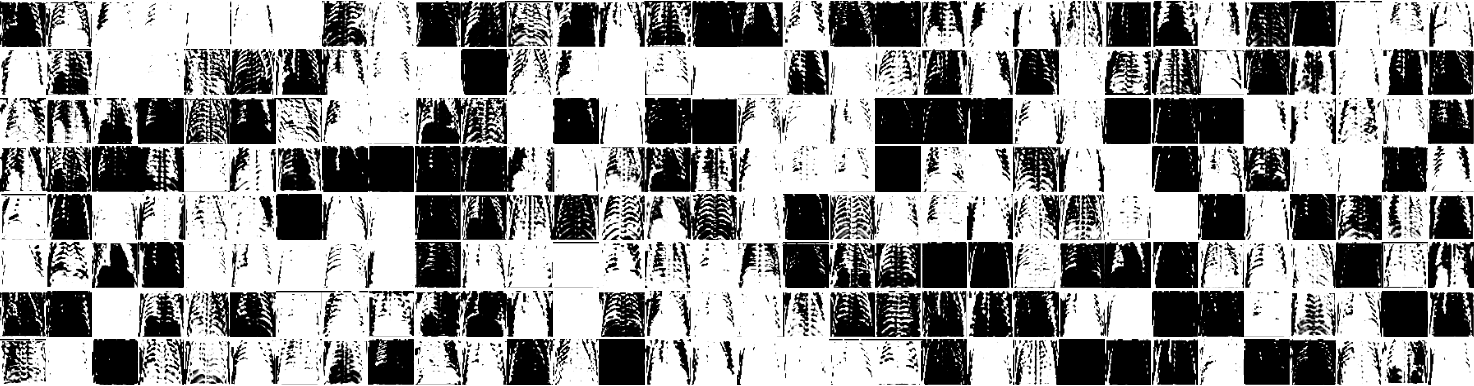}};
        \end{scope}
        \begin{scope}[cm={1,0.7,0,1,(3.7,0)}]
            \node[transform shape, draw, black, ultra thick, inner sep=0.2mm] {\includegraphics[width=.1\textwidth]{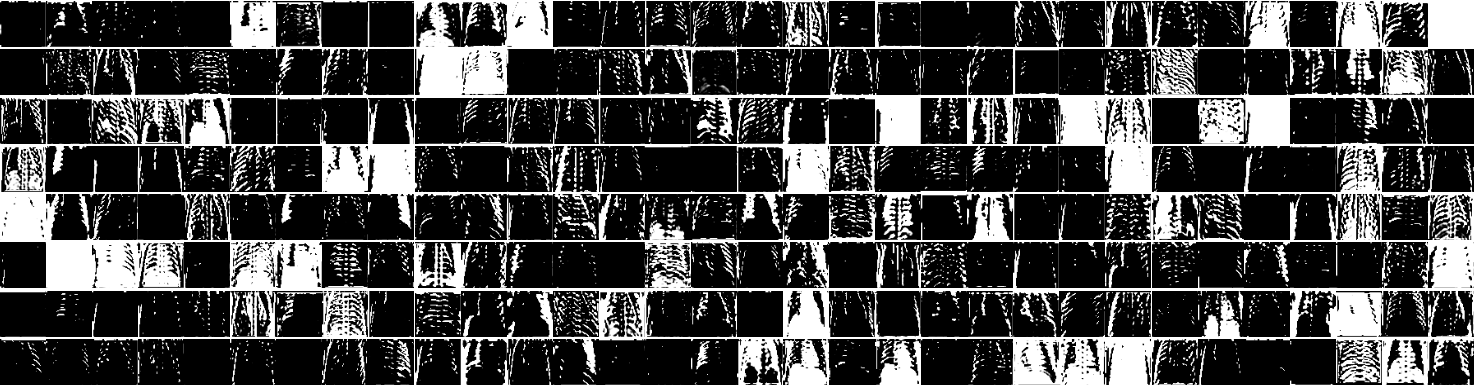}};
        \end{scope}
        
        \begin{scope}[cm={1,0.7,0,1,(4.3,0)}]
            \node[transform shape, draw, black, ultra thick, inner sep=0.2mm] {\includegraphics[width=.1\textwidth]{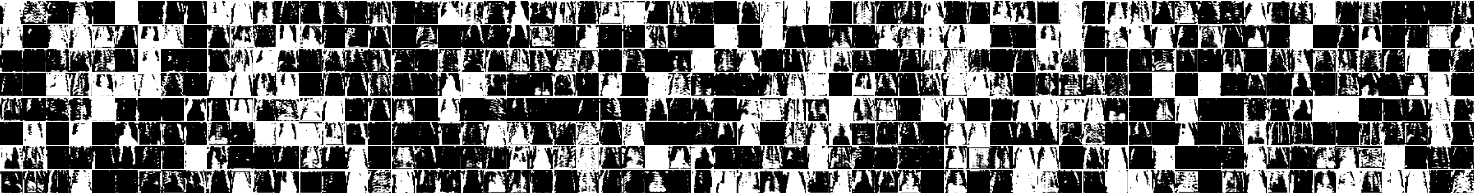}};
        \end{scope}
        \begin{scope}[cm={1,0.7,0,1,(4.5,0)}]
            \node[transform shape, draw, black, ultra thick, inner sep=0.2mm] {\includegraphics[width=.1\textwidth]{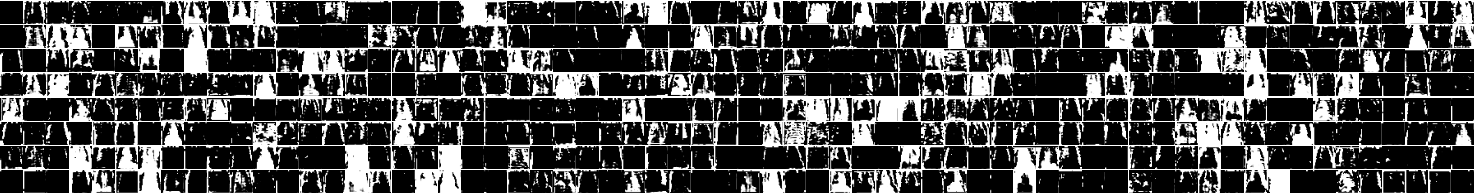}};
        \end{scope}
        \begin{scope}[cm={1,0.7,0,1,(4.7,0)}]
            \node[transform shape, draw, black, ultra thick, inner sep=0.2mm] {\includegraphics[width=.1\textwidth]{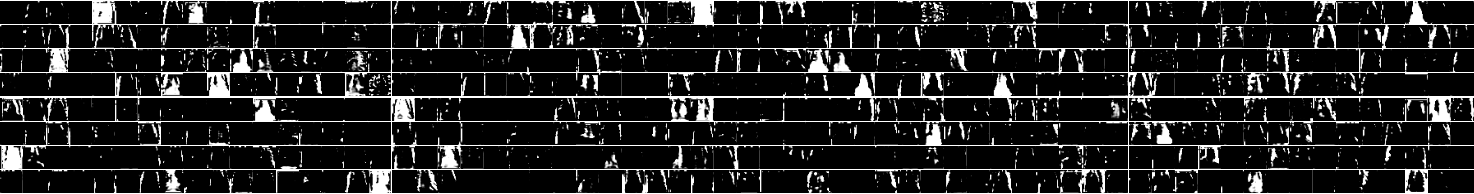}};
        \end{scope}
        
        \begin{scope}[cm={1,0.7,0,1,(5.3,0)}]
            \node[transform shape, draw, black, ultra thick, inner sep=0.2mm] {\includegraphics[width=.07\textwidth]{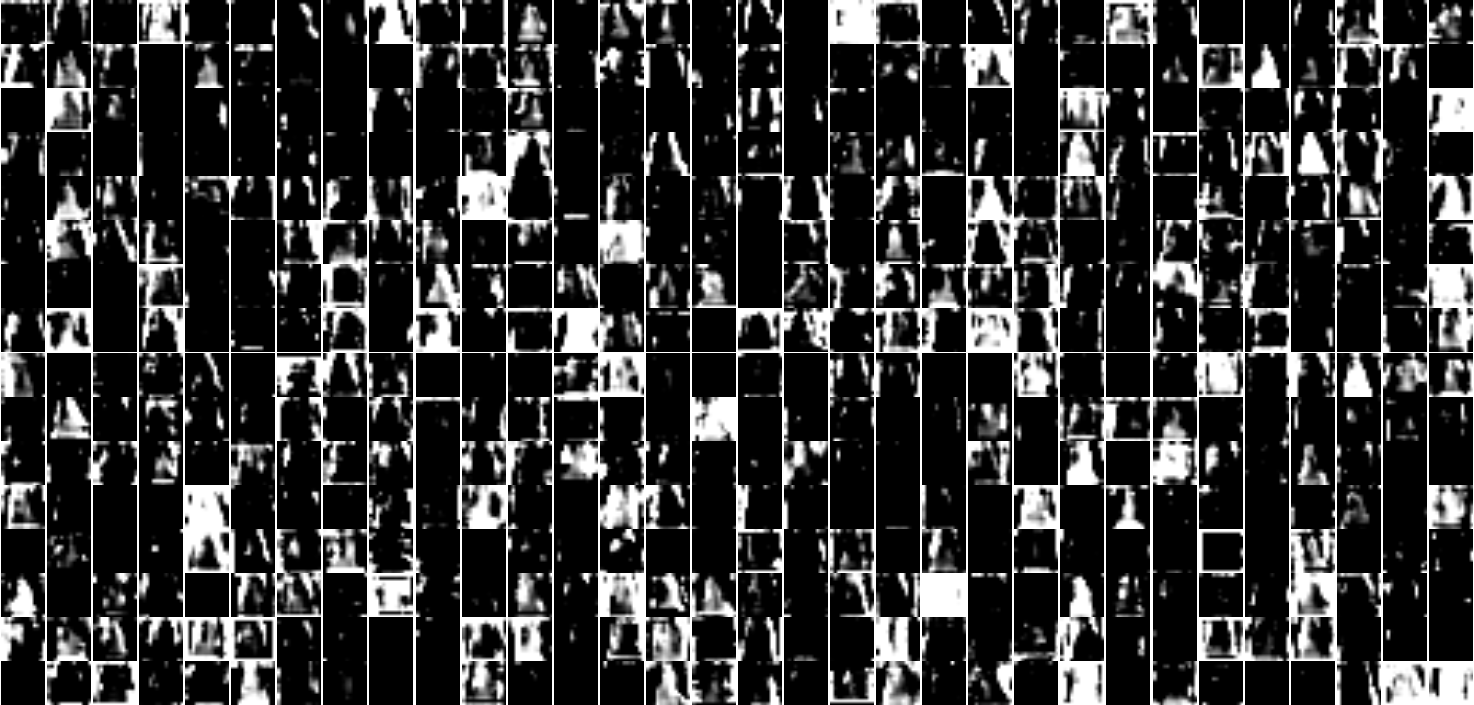}};
        \end{scope}
        \begin{scope}[cm={1,0.7,0,1,(5.5,0)}]
            \node[transform shape, draw, black, ultra thick, inner sep=0.2mm] {\includegraphics[width=.07\textwidth]{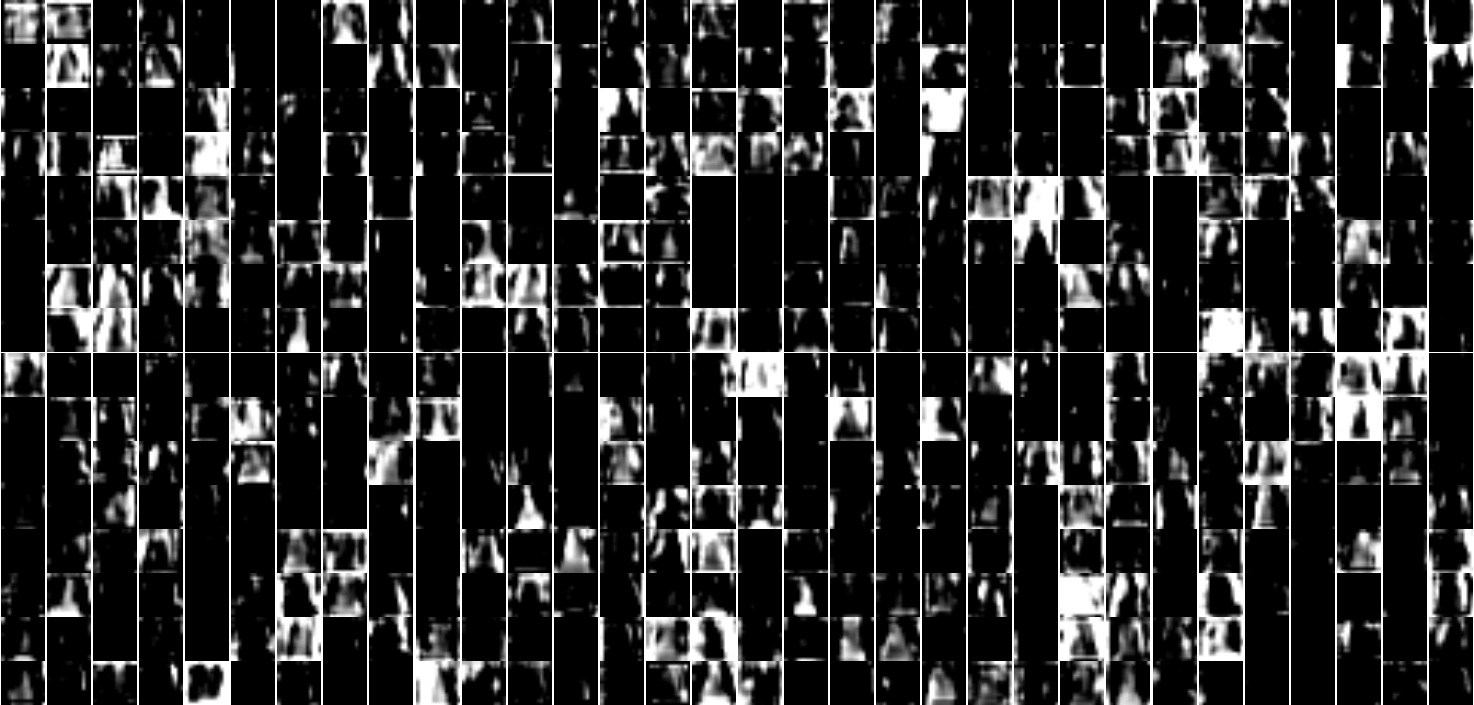}};
        \end{scope}
        \begin{scope}[cm={1,0.7,0,1,(5.7,0)}]
            \node[transform shape, inner sep=0.2mm] {\includegraphics[width=.07\textwidth]{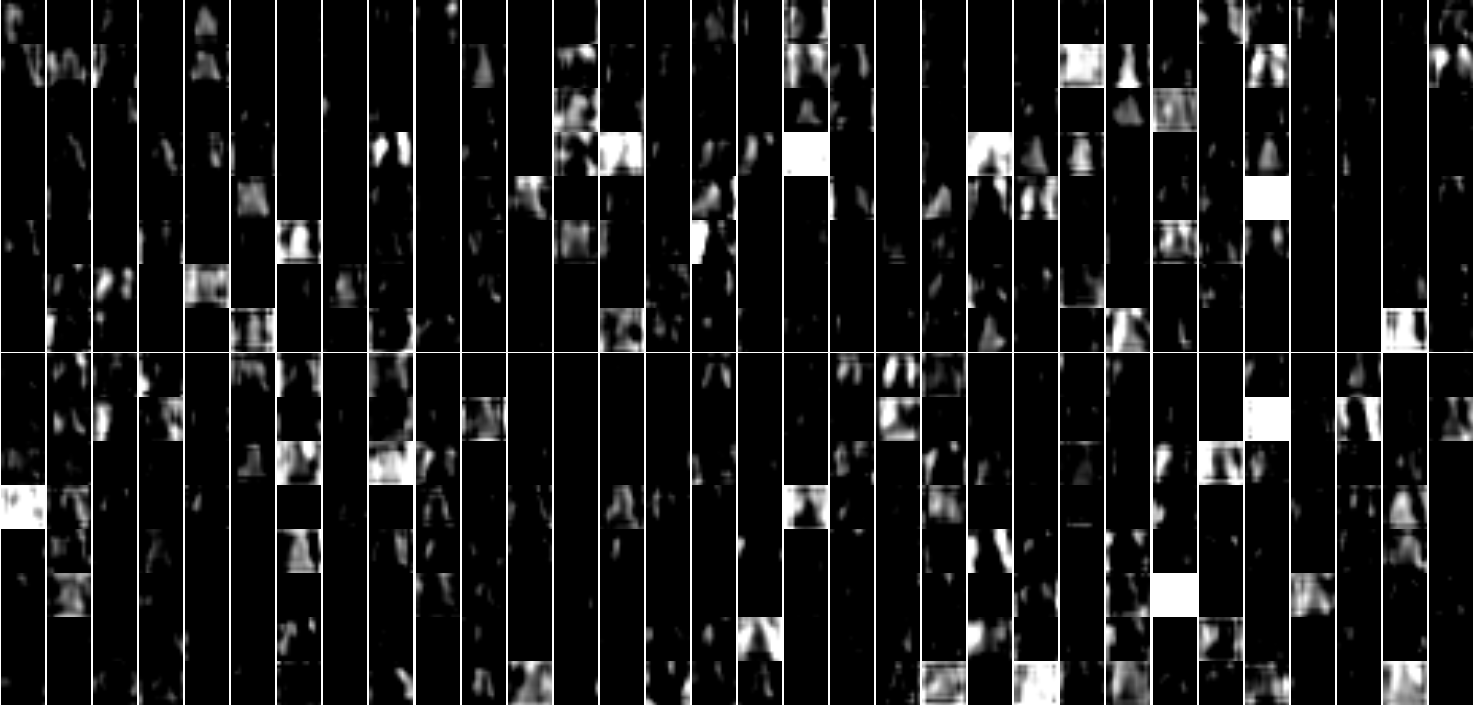}};
        \end{scope}
    \end{tikzpicture}
    \label{fig:VGG16-featuremaps-one-observation}
\end{figure}

In our work, we adapt the architecture of VGG16. The key component is feature extraction using VGG16 feature maps. We feed every image in Chest X-ray data set into the architecture of VGG16. This means that each image goes through the VGG16 procedure and we can produce many feature maps according to the parameters of the VGG16 architecture. The architecture of VGG16 model is presented in Figure \ref{fig:vgg16-architecture}. A brief overview of the filter kernels used in the VGG16 model is presented in Figure \ref{fig:VGG16-filters}. A visualization of such feature map is produced in Figure \ref{fig:VGG16-featuremaps-one-observation}.

From Figure \ref{fig:VGG16-filters}, we can see that each filter is designed with certain small patterns that aims to capture certain information in a local 3-by-3 window on an image. These designs are direct production of the work by \cite{Simonyan2014}. Though these filters produce robust performance in \cite{Simonyan2014}, any new adaption of these filters on a new dataset may or may not produce intuition to human users. \textbf{We regard the usage of many filters from a pre-trained model without any feature assessment methodology to be the major reason why transfer learning relying on a deep CNN such as VGG16 inexplainable.}

Suppose we are given image data. This means one observation is a colored picture and it is a 3-dimensional array (a tensor). If it is a black and white picture, we treat is as a 2-dimensional array. However, in order to use VGG16, it is required that the input dimension to be 224 by 224 by 3. Any black and white picture is resized into 3-dimensional tensor by treating each color to have the same greyscale. 

By using the filters (see the previous slide), we are able to extract certain information from the original image. These new features are information that are essentially some transformation of the original pixels based on filters created to detect certain patterns in another dataset. 

A deep CNN such as VGG16 typically has many convolutional layers. They are formed by standard techniques such as convolutional operation, pooling, and so on. There is not yet any explainable measure in the literature that can help us measure the explainability of these feature maps. 

\begin{figure}
    \centering
    \caption{These are the 512 feature maps extracted from the last convolutional layer of the VGG16 model. In transfer learning, we adopt a previously trained CNN model on current data set (i.e. pneumonia data). Due to lack of feature assessment procedure, we expect many of these feature maps to be inexplainable to detect the disease cases which are noisy information that do not contribute to a robust prediction performance. \textbf{The motivation is to generate a robust understanding which features can explain the prediction performance.}}
    \includegraphics[width=0.47\textwidth]{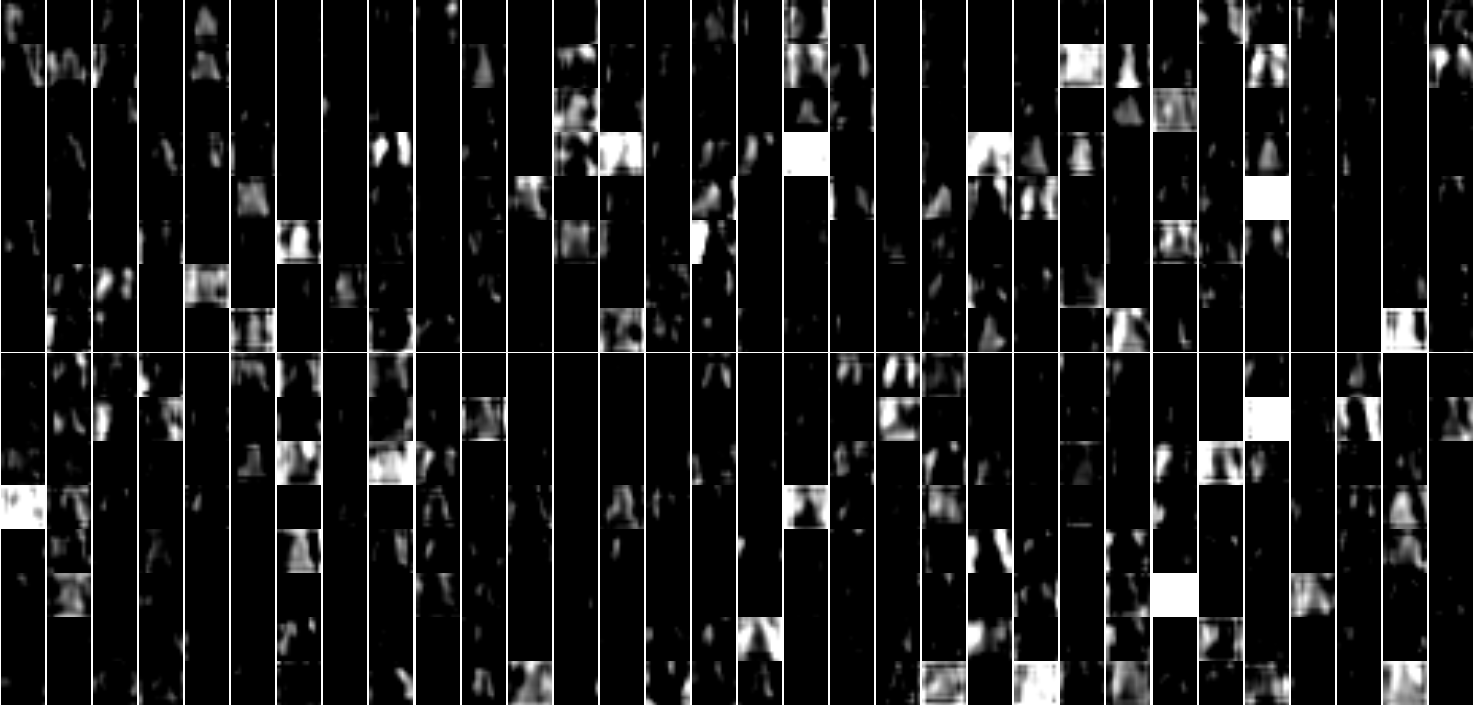}
    \label{fig:VGG16-block5-conv3}
\end{figure}

\subsection{Feature Assessment and Predictivity}

The high I-score values suggest that local information possess capability to have higher lower bounds of the predictivity. This is worth noticing because this information leads to not just high prediction performance but also explainable power. Specifically, the statistics of I-score does not rely on any assumption of the underlying model. This property eliminates the potential harm a fitted model can bring when we try to explain how features affect the prediction performance. Hence, the first condition $\mathcal{C}1$ satisfies. In addition, the magnitude of I-score allows statisticians to carry out feature assessment in order to make comparisons which subset of features are more explainable and influential at making predictions. This satisfies the second condition $\mathcal{C}2$ of an explainable measure. Third, the construction of the proposed I-score associates feature explainability with the predictivity of the features. This implies that I-score measure satisfies the third condition $\mathcal{C}3$. \textbf{Hence, the proposed statistics I-score is an explainable measure and therefore the final score is the explainability of the variables.}

\begin{figure}
    \centering
    \caption{This graph presents empirical relationship between proposed I-score and AUC. To plot this graph, we investigate the last convolutional layer (namely ``block5-conv3'') in the VGG16 architecture, i.e. there are 512 of these features extracted. We compute I-score and AUC value for each one of the 512 features.}
    \includegraphics[width=0.47\textwidth]{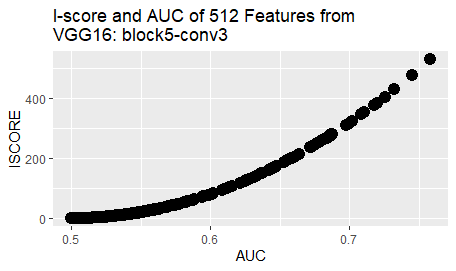}
    \label{fig:marginal-iscore-vgg16-block5-conv3}
\end{figure}

Suppose we have true label $Y$ and a predictor $\hat{Y}$. We can compute AUC of this predictor $\hat{Y}$. First, we use thresholds to convert $\hat{Y}$ into two levels, i.e. ''0''and ''1'', where the thresholds are formed by the unique levels of the real values in $\hat{Y}$. Then we compute a confusion table based on $Y$ and converted $\hat{Y}$. The confusion table gives us sensitivity and specificity which allows us to plot ROC. The area under the ROC path is Area-Under-Curve (AUC). The notion of sensitivity is interchangeable with recall or true positive rate. In a simple two-class classification problem, the goal is to investigate covariate matrix $X$ in order to produce an estimated value of $Y$. From the output of a Neural Network model, the predicted values are always between 0 and 1, which acts as a probabilistic statement to describe the chance an observation is class 1 or 0. Given a threshold between 0 and 1, we can compute sensitivity to be the following
\begin{equation}\label{eq:definition-sensitivity}
    \begin{array}{rcl}
        \text{Sensitivity} 
        &=& \mathlarger{\frac{\text{True Positive}}{\text{Positive}}} \\
        &=& \mathlarger{\frac{\text{\# of Correctly Classified Pneumonia Images}}{\text{\# of Pneumonia Images}}}
    \end{array}
\end{equation}
On the other hand, specificity is also used to create ROC curve. Given a certain threshold between 0 and 1, we can compute specificity using the following
\begin{equation}\label{eq:definition-specificity}
    \begin{array}{rcl}
        \text{Specificity}
        &=& \mathlarger{\frac{\text{True Negative}}{\text{Negative}}} \\
        &=& \mathlarger{\frac{\text{\# of Correctly Classified Non-Pneumonia Images}}{\text{\# of Non-Pneumonia Images}}}
    \end{array}
\end{equation}
Given different thresholds, a list of pair of sensitivity and specificity can be created. The Area-Under-Curve (AUC) is the area under the path plotted using pairs of sensitivity and specificity that is generated using different thresholds. For empirical results, please see Figure \ref{fig:marginal-iscore-vgg16-block5-conv3}.

\subsection{Explainability and Interpretation} % A subsection can be created just before a set of slides with a common theme to further break down your presentation into chunks

\begin{figure*}
\centering
\caption{The figure presents ten samples from diseased classes. For each image, we develop 512 features using the VGG16 architecture. However, the significantly large number of filters from a pre-trained model renders the newly constructed features inexplainable. In regarding to this major pitfall, we propose to use I-score, an explainable measure, that produces explainability of variables and describes how a subset of variables affect the prediction performance. \textbf{Among the 512 VGG16 features, the proposed statistics I-score can use the top 19 most explainable features to warn human users the location of chest X-ray that is directly associated with the pneumonia disease in patients.}}
\includegraphics[width=0.97\textwidth]{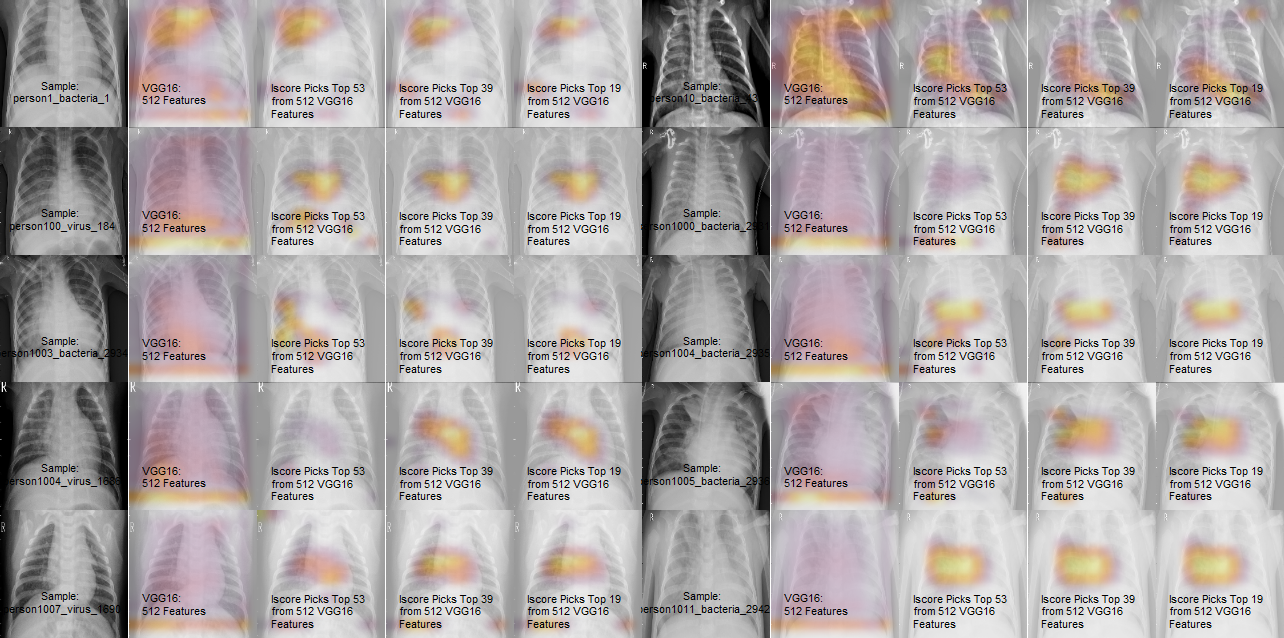} 
\label{fig:pneumonia-data-cam-iscore-visualization-diseased}
\end{figure*}

Figure \ref{fig:pneumonia-data-cam-iscore-visualization-diseased} presents ten samples from diseased classes. Each sample we present five plots: (1) the original picture sized 224 by 224, (2) visualization of the last convolutional layer of VGG16 (512 features) using heat-map generated from CAM, (3) the same visualization but using I-score to pick the top 53 out of 512 features from the last convolutional layer of VGG16 (I-score greater than 200), (4) top 39 out of 512 features (I-score greater than 150), and (5) top 19 out of 512 features (I-score greater than 100). 

In transfer learning, we adopt a previously trained model such as VGG16 and apply the convolutional features on chest x-ray data set. Due to significantly large number of filters in previous model and lack of feature selection method, it is extremely challenging to see the location in the image that is used to generate good performance using all 512 convolutional features which render the project inexplicable. 

The proposed method I-score is capable of \textbf{selecting and explaining}, out of 512 convolutional features from a previously trained model, the important features (sometimes as little as less than 30 features) to create explainability and interpretability and the proposed procedure uses heat-map to suggest the exact location that the disease may occur in patients.
    
\begin{table*}[!t]
\centering
\caption{The table presents test set performance of the Pneumonia Chest X-ray Detection data set. The performance of held-out test set is measured by the Area-Under-Curve (AUC) from Receiver Operating Characteristic (ROC). The proposed approach works with images that are resized into 224 by 224 by 3 (i.e. 150,528 pixels) and the procedure uses transfer learning to take a combination of pre-trained models such as VGG16 to generate 512 features using its deep convolutional layers. Because models like VGG16 are trained using a different set of image data and with hundreds of filters tuned from prior learning experience, there is little hope that these features grant us passages explain exactly how the explanatory variables influence the response variable which raises ambiguity for explainability and interpretability. The proposal here is to use I-score. From using I-score and Backward Dropping Algorithm, we are able to select up to 400 features to construct a neural network with less than 20,000 parameters to produce state-of-the-art test set performance at 99.7\% while originally the 512 features directly from VGG16 might need 1.31 million parameters to produce equal results. This is a 98\% dimension reduction on the number of parameters required for building neural network architecture. Although in our experiments, we used seven well-known deep CNN models, the proposed method can be generalized and adapted to extract informative, explainable, and interpretable features (or variables) from the last convolutional layer generated by any combination deep CNN models.}
\begin{tabular}{llcc}
\toprule
Method & No. of Param. & Test AUC \\
\hline \hline
Previous: & \\
VGG16 \citep{Ayan2019} & approx. 138 million & 87.0\% \\
\cite{Saraiva2019} & & 94.5\% \\
AlexNet \cite{Togacar2019} & approx. 60 million & 96.8\% \\
\cite{Cohen2019} & & 99.0\% \\
\cite{Rajaraman2018} & approx. 0.8 - 40 million & 99.0\% \\
\hline
Proposed: \\
Interaction-based Feature Learning / Feature Generation \& Extraction \\
224 by 224 by 3 \\
(use deep convolutional layers) \\
512 Features from VGG16 + NeuralNet (Single-layer) & $>$ 131,000 & 98.7\% \\
512 Features from VGG19 + NeuralNet (Single-layer) & $>$ 131,000 & 98.8\% \\
512 Features use I-score: \\
$\sim$ Top 0.5\% Modules (approx. 70 features) & approx. 8,000 & 98.3\% \\
$\sim$ Top 1\% Modules (approx. 100 features) &  approx. 9,000 & 98.6\% \\
$\sim$ Top 2.5\% Modules (approx. 200 features) &  approx. 10,000 & 99.3\% \\
$\quad$ removal of the above 200 features & & 57.4\% \\
5,216 Features from 7 different CNNs + NeuralNet (Single-layer) & $>$ 1.31 million & 99.7\% \\
5,216 Features use I-score: \\
$\sim$ Top 1\% Modules (approx. 100 features) &  approx. 13,000 & 98.8\% \\
$\sim$ Top 2\% Modules (approx. 150 features) & approx. 19,000 & 99.2\% \\
$\sim$ Top 5\% Modules (approx. 400 features) & approx. 20,000 & 99.7\% \\
$\quad$ removal of the above 400 features & & 54.2\% \\
\bottomrule
\end{tabular}
\label{tab:pneumonia-lab-results}
\end{table*}

To further confirm the improvement that the proposed method delivered, we conduct experiments by removing the influential features by I-score and we present AUC with the data (where these important features are not presented). For example, in Table \ref{tab:pneumonia-lab-results} of the proposed approach for the top 2.5\% modules, we delivered 99.3\% AUC with approximately 200 features. The removal of these features drops AUC from 99.3\% to 57.4\% which is almost close to random guessing. For the top 5\% of modules of 5,216 features from 7 different CNNs, we delivered 99.7\% using approximately 400 features which are selected by I-score from the last convolutional layer of these ultra-deep CNNs. The removal of these features drop AUC to 54.2\%. 

\section{Conclusion}
First, the paper provides a novel and rigorous definition for explainable and interpretable feature assessment and selection methodology (please see the boldface definition for three major conditions $\mathcal{C}1$, $\mathcal{C}2$, and $\mathcal{C}3$ for fulfilling the necessary conditions of explainable and interpretable feature selection method). Under this guidance, we define the explainability and interpretablility of a set of variables to be the final importance score measured and evaluated by only explainable and interpretable feature assessment and selection methodology. This allows us to establish rigorous discussion on the explainability and interpretability of features and variables specifically when it comes to explain what is the influence and impact a set of variables have on response variable.

Next, this paper delivers a novel interaction-based methodology to interpret and explain deep learning models while maintaining high prediction performance. In addition, we provide a way to contribute to the major issues about explainability, interpretability, and trustworthiness brought up by DARPA. We introduce and implement a non-parametric and interaction-based feature selection methodology. Under this paradigm, we propose \emph{Interaction-based Feature Learning} that heavily relies on using an explainable measure, I-score, to evaluate and select the explainable features that are created from deep convolutional neural networks. This approach learns from the final convolutional layer of any deep CNN or any combination of deep learning frameworks. For example, we use the proposed method I-score to assess and select the explainable and interpretable features from many well-known deep CNN models including VGG16, VGG19, DenseNet121, and so on to form the input layers for neural network. The proposed methods have been presented with both artificial examples and real data application in the Pneumonia X-ray image data set. We conclude from both simulation and empirical application results that I-score shows unparalleled potential to explain informative and influential local information in a variety of large-scale data sets. The high I-score values suggest that local information possess capability to have higher lower bounds of the predictivity. This is worth noticing because this information leads to not just high prediction performance but also explainable power. The proposed methodology \emph{Interaction-based Feature Learning} rely heavily on using I-score to select, combine, and explain features for neural network classifiers. This establishes a domain of using I-score with neural network (as well as with Backward Dropping Algorithm) which we regard as the field of the undiscovered field of Interaction-based Neural Network. Although we only point out two approaches in this paper, we do believe in the future there will be many other approaches within the field of Interaction-based Neural Network. We encourage both the statistics and computer science community to further investigate this area to deliver more transparency and trustworthiness to deep learning era and to build a world with truly Responsible A.I.. 

\section{Acknowledgments}
We would like to dedicate this to H. Chernoff, a well-known statistician and a mathematician worldwide, in honor of his 98th birthday and his contributions in Influence Score (I-score) and the Backward Dropping Algorithm (BDA). We are particularly fortunate in receiving many useful comments from him. Moreover, we are very grateful for his guidance on how I-score plays a fundamental role that measures the potential ability to use a small group of explanatory variables for classification which leads to much broader impact in fields of pattern recognition, computer vision, and representation learning.

\section{Code}
Code for data cleaning and analysis is provided upon requests as part of the proposed statistical package. It will be uploaded to this [Github Repository] once the paper has been conditionally accepted.

\section{Disclosure of Funding}
This research is supported by National Science Foundation BIGDATA IIS 1741191.

%%%%%%%% REFERENCE PAGE STARTS HERE %%%%%%%%
%\clearpage
\bibliographystyle{aaai}
\bibliography{refs}

\newpage
\section*{Supplement}

\label{supplement}In this section, we illustrate proposed methodologies on artificial examples. We introduce four artificial examples. These simulated examples demonstrate the usage of I-score and Backward Dropping Algorithm in a variety of different practical environments including both discrete and continuous covariates with a number of pairs of $n/p$ ratio while $n$ is the number of observations in training data and $p$ is the number of explanatory variables in training data.  The first example starts with a two-module setup with explanatory variables to be drawn from $\text{Bernoulli}(1/2)$ which represents the key issues of how to use I-score and Backward Dropping Algorithm. In the second example, we raise the complexity of the design of the response variable by introducing four modules. Though the computation of I-score relies on partition retention, we use the third example to illustrate the advantage of I-score can be well exploited using discretized partition. We land on a fourth example of which we use a model that is unconventional to the first three by injecting $\exp(\cdot)$ function into the model. Without the use of I-score, we do not have advantage to gain (or improve) on prediction performance. If I-score is adopted, one can gain huge improvement for releasing the potential information for a data set.

\subsection{Artificial Example I: Two Modules with Discrete Variables}
To illustrate some of our key issues, consider a small artificial example. Suppose we have a data set with 50 variables, $\{X_i: 1 \le i \le 50\}$, and $X_i \sim \text{Bernoulli}(1/2)$, each is drawn independently. This means each $X_i$ is binary and may take values 0 and 1, each with probability 1/2. Suppose an observed variable $Y$ is defined by the following equation, model \ref{eq:artificial_example_1},
\begin{equation}\label{eq:artificial_example_1}
Y = \left\{
\begin{array}{ll}
X_1 + X_2 & (\text{mod } 2) \text{ w/ prob. } 0.5 \\
X_2 + X_3 + X_4 & (\text{mod } 2) \text{ w/ prob. } 0.5 \\
\end{array}
\right.
\end{equation}
where $X_1$, $X_2$, $X_3$, and $X_4$ are 4 influential variables out of 50 observed variables. The rest 46 observed variables are noisy variables that do not contribute information to the response variable $Y$. In this case, the knowledge of $X_1$ itself is not enough for us to predict response variable $Y$. For example, if a method successfully identifies $X_1$ as one of the influential variable, the best that this variable, $X_1$, can do is 50\% accuracy. We assume $X_i$ were selected independently to be 1 with probability 0.5. Hence, none of the individual $X_i$ has a marginal effect on $Y$. 

Suppose we know the model and we want to compute which variable sets are predictive of $Y$, and how predictive, when $\textbf{X} = (X_1, X_2, ..., X_{50})$ is given. From the definition of underlying model in this artificial example (\ref{eq:artificial_example_1}), it is obvious that there are two clusters of variable sets $S_1 = \{X_1, X_2\}$ and $S_2 = \{X_2, X_3, X_4\}$ that are potentially useful in the prediction. We treat the highest correct prediction rate possible for a given variable set as an important parameter and call this predictivity $(\theta_c)$. In this case, we can compute the predictivity for $S_1$ as $\theta_c(S_1) = 75\%$. The predictivity for $S_2$ is $\theta_c (S_2) = 75\%$ as well. Incidentally, the predictivity of the union of $S_1$ and $S_2$ is also 75\%.

From above discussion, the realization is that by using variable sets $S_1$ and $S_2$ one can predict response variable $Y$ correctly 75\% of the time. This is beyond doubt, because upon observing $\textbf{X} = (X_1, ..., X_{50})$ we can predict
\begin{equation}\label{eq:artificial_example_1_one_module}
\hat{Y} = X_1 + X_2 \text{ (mod } 2) \text{ with probability } 1/2
\end{equation}
and the other $1/2$ of time it is random guessing so we are hitting 50\% of the 50\% which is 25\%. Putting the above together, we can formerly write the theoretical prediction rate of a correct model specification of module $\{X_1, X_2\}$ is the following
\begin{equation}\label{eq:artificial_example_1_one_module_bayesrate}
\begin{array}{rcl}
\theta_c(S_1) 
&=& \theta_c([X_1,X_2]) \\
&=& \frac{1}{n} \sum_i \mathds{1}(\hat{Y} = Y) \\
&=& \frac{1}{n} \sum_i \mathds{1}(\underbrace{(X_1 + X_2)}_\text{(mod 2)} = Y) \\
&=& 50\% + 50\% \cdot 50\% \\
&=& 75\% \\
\end{array}
\end{equation}
while in the third line we plug in the predictor $\hat{Y}$ that is defined using the module in equation \ref{eq:artificial_example_1_one_module}. For simplicity, we use the formula $\frac{1}{n} \sum_i (\hat{Y} = Y)$ to measure the prediction rate of the predictor $\hat{Y}$ constructed using variable set $S_1 = \{X_1, X_2\}$. A more formal version is the predictivity defined in \citep{lochernoffzhenglo2015} and \citep{lochernoffzhenglo2016}. The notion of predictivity (or correct theoretical prediction rate) $\theta_c$ on a variable set $X$ is defined as $\theta_c[p_{X_d}, p_{X_u}] = \frac{1}{2} \sum_{x \in \Pi_X} \max\{p_{X_d}(x), p_{X_u}(x)\}$ while the subscript $d$ indicates case group and $u$ indicates control group. The partition $\Pi_X$ is defined using $X$. For example, if the set is $\{X_1, X_2\}$ with each variable to be dichotomous, then the partition has size $2^2 = 4$.

It is not difficult to show that the strategy of predicting with $S_1$ returns prediction accuracy of 75\% by expectation which is also the highest percent accuracy $S_1$ can theoretically achieve. The same result goes with $S_2$ as well. 

Alternatively, suppose the statistician has no knowledge of the underlying model. The conventional procedure is to use all the features in the data to make predictions. We demonstrate the idea using the following experiment. We create 50 variables with sample $n$ drawn randomly from Bernoulli distribution. This sample size $n$ in training vary from 50, 100, or 1000. We define the underlying model (response variable) $Y$ based on model (\ref{eq:artificial_example_1}). We treat this data set with sample size $n$ to be our training set. We will run common machine learning algorithm as well as proposed algorithm on training set and test the results on newly generated $1000$ test data points. These common machine learning algorithms are Bagging, Logistic, Random Forest (RF), iterative Random Forest (iRF), Neural Network (NN). We then set a different seed randomly and repeat the above experiment 10 times. In the end, we take the average accuracy of test set performance. The out-of-sample test set performance is measured using Area-Under-Curve (AUC) from Receiver Operating Characteristic (ROC). For each value of training sample size, we start by making prediction using all variables (including noisy variables). To demonstrate the importance of variable selection, we also use I-score to select the important variables and take the first module to make predictions. The results are compared in the following. In the following table (see Table \ref{tab:simulation-discrete-easy-fix-p-change-n}), we present the test set performance for major machine learning algorithm and proposed method. It is obvious that lack of sample size can be a challenge for machine learning algorithms such as Bagging, Logistic, Random Forest (RF), iterative Random Forest (iRF), and Neural Network (NN). However, proposed method, I-score, does not subject to this problem as much as other algorithms. The results are shown in the following graph (see Table \ref{tab:simulation-discrete-easy-fix-p-change-n}).

\begin{table*}[!t]
\centering
\caption{This table presents simulation results for model \ref{eq:artificial_example_1}. The theoretical prediction rate is calculated in equations \ref{eq:artificial_example_1_one_module_bayesrate}, which is at 75\%. In other words, we expect prediction performance on out-of-sample test set to be approximately 75\% on average. For each experiment below, the out-of-sample test set has 1000 sample data points and the performance is calculated using Area-Under-Curve (AUC) from Receiver Operating Characteristic (ROC). Each experiment we change in-sample training size to be 50, 100, or 1000. We fix all data to have number of variables to be 50. Without using the proposed method, it is nearly impossible to select the influential variables that help us explain the response variable. In this case, algorithm would apply to all 50 variables in the data including noisy and non-informative variables. Because of substantial amount of noisy variables are included in the model fitting procedure, common classifiers produce poor results around 50\% AUC on held-out test set and this performance is no different random guessing. For Neural Network (NN), it is challenging even for 1000 samples. However, prediction can be drastically improved using I-score to select the explainable features. In this case, we are able to use I-score to select the module $\{X_1, X_2\}$. This is the most important module because it has the highest I-score value. Alternative to using all variables, the proposed method suggests to use only the top module $\{X_1, X_2\}$. This top module $\{X_1, X_2\}$ is a result from the Backward Dropping Algorithms with the highest I-score values and is used in different algorithms in this table. These improvements provide major contribution to predictivity and explainability to complicated and large-scale data set.}
\begin{tabular}{clccccc}
\toprule
Variables: 50	&		&	Test AUC	\\
Training Sample Size:	&	Algorithms	&	Bagging	&	Logistic	&	RF	&	iRF	&	NN	\\
\hline
50	&	All Var.	&	0.51	&	0.51	&	0.50	&	0.50	&	0.52	\\
	&	I-score: Top Mod.	&	0.65	&	0.65	&	0.65	&	0.64	&	0.52	\\
\hline
100	&	All Var.	&	0.54	&	0.52	&	0.50	&	0.52	&	0.51	\\
	&	I-score: Top Mod.	&	0.74	&	0.74	&	0.75	&	0.75	&	0.51	\\
\hline
1000	&	All Var.	&	0.62	&	0.50	&	0.54	&	0.63	&	0.52	\\
	&	I-score: Top Mod.	&	0.75	&	0.72	&	0.75	&	0.73	&	0.75	\\
\bottomrule
\end{tabular}
\label{tab:simulation-discrete-easy-fix-p-change-n}
\end{table*}

\begin{table*}[!t]
\centering
\caption{This table presents simulation results for model \ref{eq:artificial_example_1}. The theoretical prediction rate is calculated in equations \ref{eq:artificial_example_1_one_module_bayesrate}, which is at 75\%. In other words, we expect prediction performance on out-of-sample test set to be approximately 75\% on average. For each experiment below, the out-of-sample test set has 1000 sample data points and the performance is calculated using Area-Under-Curve (AUC) from Receiver Operating Characteristic (ROC). Continuing from table \ref{tab:simulation-discrete-easy-fix-p-change-n}, we fix in-sample training size to be 1000 data points and we allow number of variables in the toy data to be 100 and 200. From 50 variables, this is a more challenging situation because it lower the chance of searching for the correct variable modules.}
\begin{tabular}{clccccccc}
\toprule
Training Sample Size: 1000	&		&	Test AUC	\\
Variables:	&	Algorithms	&	Bagging	&	Logistic	&	RF	&	iRF	&	NN	\\
\hline
100	&	All Var.	&	0.57	&	0.52	&	0.51	&	0.54	&	0.51 \\
	&	I-score: Top Mod.	&	0.74	&	0.72	&	0.74	&	0.73	&	0.75 \\
\hline
200	&	All Var.	&	0.51	&	0.52	&	0.51	&	0.51	&	0.52 \\
	&	I-score: Top Mod.	&	0.74	&	0.72	&	0.74	&	0.74	&	0.74 \\
\bottomrule
\end{tabular}
\label{tab:simulation-discrete-easy-fix-n-change-p}
\end{table*}

The purpose for the above simulation results (see Table \ref{tab:simulation-discrete-easy-fix-p-change-n}) is to demonstrate the importance of successfully identifying the influential variables among many noisy variables. The data set has 50 variables out of which only the first four variables are important. We allow in-sample training data size to vary in 50, 100, or 1000. Each value of the in-sample training size, we conduct experiments using all 50 variables and we also run the same experiments using variables selected by I-score. Each experiment we run a selection of machine learning algorithms: Bagging, Logistic, Random Forest (RF), iterative Random Forest (iRF), and Neural Network (NN). For the first four algorithms, we use default parameter values. For Neural Network (NN), we use a one-layer neural network with 30 nodes, a number sufficient to analyze a handful of important variables based on tuning performance.\footnote{This is a result from tuning, but we fix the architecture before and after we use I-score.} After training from each experiment, we test the performance on out-of-sample test set that has 1000 data points. We produce the Area-Under-Curve (AUC) from Receiver Operating Characteristic (ROC). For AUC, a random guessing of 0's and 1's will have 50\% while a perfect prediction will hit 100\% of AUC. From the underlying model defined in model \ref{eq:artificial_example_1}, we compute 75\% of performance which should give us AUC of approximately 75\% on average given the correct variables specified. From Table \ref{tab:simulation-discrete-easy-fix-p-change-n}, we observe that the performance of all methods are relatively poor when we only allow 50 data points in the training set. This lack of training data can be observed from test set AUC values. However, though not able to hit theoretical prediction rate of 75\%, I-score can still pick up some signal to push almost all algorithms up from around 50\% performance. When we increase in-sample training sample size to 100, we immediately see that almost all methods start to hit the theoretical performance rate of 75\%. Amongst these methods, Neural Network (NN) still require ample amount of data to be able to perform well. When we increase training set sample size to 1000, we start to see all methods to perform as well as theoretical prediction rate with variables selected by I-score. However, without I-score, the prediction rate on out-of-sample test set have been overall quite poor even with sufficient amount of data. 

From results in Table \ref{tab:simulation-discrete-easy-fix-n-change-p} above, we fix in-sample training size to be 1000 and we allow number of variables to be 100 and 200. From a data with only 50 variables, this is a more challenging task because this increases the difficulty of searching for the correct variable module. Although the level of difficulty is raised from previous simulation, we observe that as long as we have correct variables selected by using proposed method I-score, we are able to successfully achieve theoretical prediction rate at 75\%. This is not true if we do not conduct feature selection. The amount of noisy variables presented in this simulation washes away the signal from the data that can be used to make good predictions. This can be shown from looking at out-of-sample prediction set AUC which are generally poor given the amount of noisy variables in the data.

\subsection{Artificial Example II: Complex Form}

The above example is a standard two-module simulation environment similar to \cite{chernoffetal2009}, \cite{lochernoffzhenglo2015}, and \cite{lochernoffzhenglo2016}. To further explore how I-score and AUC behaves in an model with noisy information. We create the following simple simulation with a more complex form than the previous example. Let us consider the following equation, model \ref{eq:artificial_example_4}, to be defined as
\begin{equation}\label{eq:artificial_example_4}
    Y = \mathbb{1}\bigg(\frac{X_1 \cdot X_2}{\exp(X_3 \cdot X_4)} > 0\bigg)
\end{equation}
while $X_j$ for $j = 1, 2, 3, 4$ are dichotomous.

Suppose the statistician knows the model and the important variables. The theoretical performance would be 100\%. In other words, he knows exactly what $Y$ is and he would have guessed it correctly for all observations given to him. For simplicity, let us consider a toy data that has 1000 data points in training set and all variables are drawn from $\text{Bernoulli}(1/2)$ independently. Given $Y$ that is defined using model \ref{eq:artificial_example_4}, we can compute empirically the AUC of using $X_1$ and $X_2$ alone as a predictor that are 84\% and 83\%, respectively. The I-score of $X_1$ and $X_2$ independently are 168 and 166, respectively. The two variables together form a strong variable module. We can compute the AUC and I-score of $X_1 + X_2$ as a predictor to be 100\% and 292, respectively. We can also compute similar statistics for $X_3$ and $X_4$. Though the variables $X_3$ and $X_4$ are in the model, they are used as scaling factor and actually do not harm prediction performance if they are not included. This is because the $X_3 \cdot X_4$ takes values of 1 and 0. The exponential function of 1 and 0 produce positive numbers, i.e. $\exp(1)$ and $\exp(0)$ are both positive. Hence, these values will not affect the result of outcome variable $Y$ and only $X_1$ and $X_2$ are truly predictive and explainable. Indeed, we observe from both AUC values and I-score magnitudes that the variable set $\{X_1, X_2\}$ is the major component of the underlying model, eq \ref{eq:artificial_example_4} which implies that these variables play the real role of building the outcome variable $Y$. In other words, the variables $X_1$ and $X_2$ are the really the explainable feature with highly predictive power at explaining and interpreting the outcome variable $Y$. These statistics can be summarized in the Table \ref{tab:artificial_example_4_theoretical_rate}.

Mathematically, we can draw $X_i \sim \text{Bernoulli}(1/2)$ while $i = 1, 2, 3, 4$. In this case, we have $X_1\cdot X_2$ to have probability mass function $\mathbb{P}(X_1X_2 = 1) = 1/4$ and $\mathbb{P}(X_1X_2 = 0) = 3/4$ while $\mathbb{P}(X_1X_2) = 0$ elsewhere. In other words, the support of $X1 \cdot X_2$ is $R_{X_1 \cdot X_2} = \{0, 1\}$. Notice that $X_3$ and $X_4$ are also dichotomous. Hence, $X_3 \cdot X_4$ can only take values 0 or 1. Since $\exp(0) = 1$ and $\exp(1) \approx 2.7$, we can write the support for $X_1 \cdot X_2 / \exp(X_3 X_4)$ would be $R_{X_1 \cdot X_2 / \exp(X_3 X_4)} = \{0, 1/\exp(1), 1\}$. Then the indicator function forces the support $R_{X_1 \cdot X_2 / \exp(X_3 X_4)}$ to be back to 0 and 1, i.e. $R_{\mathbb{1}(X_1 \cdot X_2 / \exp(X_3 X_4))} = \{0, 1\}$. This renders the information of $X_3 \cdot X_4$ non-informative in this model. The reason is the following. When one of $X_1$ and $X_2$ is zero, the product of $X_1 \cdot X_2$ would be zero. This is the scenario when $Y$ takes the value zero. Alternatively, when both $X_1$ and $X_2$ take value one, the product of them would be one. This gives us $Y = 1$. The denominator of $\exp(X_3 \cdot X_4)$ gives us $1/\exp(0)$ or $1/\exp(1)$, but these values are converted to one after the indicator function. In other words, we conclude that $Y = 1$ only when $X_1 = X_2 = 1$. In any other situation from the support of $X_1 \cdot X_2$, we would have $Y = 0$. In summary, we would only need $X_1 \cdot X_2$ to make perfect predictions of the model \ref{eq:artificial_example_4} even though $X_3$ and $X_4$ are also present in the model.

% X1X2 = 0 or 1
% exp() = exp(0)=1 or exp(1)=2.7
%0 1 or 1/2.7

In reality, the statistician rarely knows the underlying model and we suppose we do not have the important features either. In this scenario, the statistician can directly start with all the variables and seek to build classifier. Alternatively, we recommend to use the proposed methods. In other words, we have the explanatory variables $X$ and response variable $Y$ which allow us to start to run the Backward Dropping Algorithm. 

\begin{table*}[!t]
    \centering
    \caption{This table presents the variable of model \ref{eq:artificial_example_4} independently. Suppose the statistician knows the model. This means that this statistician is aware of the variables used and how to use explanatory variables to construct response variable. Thus, the theoretical prediction rate is 100\%, which can be seen from row 10 in the table below (since row 10 assumes we know the specific form of the true model). The values in the table present empirical results of each component of the model \ref{eq:artificial_example_4} on a toy data of 1000 sample size (with each of the 10 experiments, row 1-10, repeated 50 times). We report the average AUC and I-score values below. It is worth mentioning that the variables $X_3$ and $X_4$ deliver very little predictive power to the formulation of response variable (or outcome variable) even though variables $X_3$ and $X_4$ are both in the model. This is because the result of an exponential function of 1 or 0 is a positive number and this does not change the result of indicator function $\mathbb{1}(\cdot)$. In other words, the values of $\exp(X_3X_4)$ can be $\exp(1)$ or $\exp(0)$. Dividing the numerator $X_1 \cdot X_2$ by either $\exp(1)$ or $\exp(0)$ will not change the sign of $X_1\cdot X_2$ and hence it would not affect the results of outcome variable $Y$. Indeed, we observe that the AUC values and the proposed I-score are both low for predictors using $X_3$ and $X_4$ (see row 3, 4, 7, and 8) which indicating very little predictive power and explainability on outcome variable $Y$. Thus, the response variable is completely determined by the variable set $\{X_1, X_2$\} which can both be selected by the proposed method, I-score, due to extremely large I-score values that indicating the high explainability and predictive power this variable set has at explaining outcome variable $Y$. We can see that I-score is tightly related to the values of AUC because I-score is directly associated with the predictivity of the variable set (this is also defined in the third condition $\mathcal{C}3$ of the explainable and interpretable feature assessment and selection method in \S2, one can also refer to predictivity in \citep{lochernoffzhenglo2015} and \citep{lochernoffzhenglo2016}).}
    \begin{tabular}{cccc}
        \toprule
        Row & Predictor & Ave. AUC & Ave. I-score \\
        \hline
        1	&	$X_1$					&	0.84	&	168.80	\\
        2	&	$X_2$					&	0.83	&	166.31	\\
        3	&	$X_3$					&	0.51	&	0.61	\\
        4	&	$X_4$					&	0.51	&	0.42	\\
        5	&	$X_1 + X_2$			&	1.00	&	292.10	\\
        6	&	$X_1 \cdot X_2$			&	1.00	&	375.70	\\
        7	&	$X_3 + X_4$			&	0.50	&	0.59	\\
        8	&	$X_3 \cdot X_4$			&	0.50	&	0.34	\\
        9	&	$\frac{X_1 X_2}{\exp(X_3 X_4)}$			&	1.00	&	306.41	\\
        10	&	$\mathbb{1}(\frac{X_1 X_2}{\exp(X_3	X_4)}	>	0)$	&	1.00	&	375.70	\\
        \bottomrule
    \end{tabular}
    \label{tab:artificial_example_4_theoretical_rate}
\end{table*}

\begin{table*}[!t]
    \centering
    \caption{This table presents simulation results of the test set performance. In this simulation, we use model \ref{eq:artificial_example_4}. There are two panels. In Panel A, we fix number of variables in training set to be 50. We allow in-sample training set size to vary in 50, 100, or 1000. We fix the out-of-sample test set to be 1000 sample size throughout this experiment. In Panel B, we fix in-sample training size to be 100 and we allow number of variables to be 100 or 200. We observe that for smaller sample size and common machine learning algorithms produce poor results without using I-score. The performance can be significantly improved with the aid of the proposed method at any pair of sample size and number of variables.}
    \begin{tabular}{llcccc}
        \toprule
        Panel A \\
        Variables: 50	&		&	Test AUC					\\
        Training Sample Size:	&		&	Logistic	&	RF	&	NN	\\
        \hline
        50	&	All Var.	&	0.51	&	0.61	&	0.57	\\
        	&	I-score: Top Mod.	&	0.80	&	0.80	&	0.70	\\
        \hline
        100	&	All Var.	&	0.98	&	0.82	&	0.52	\\
        	&	I-score: Top Mod.	&	1.00	&	1.00	&	1.00	\\
        \hline
        1000	&	All Var.	&	1.00	&	0.94	&	0.84	\\
        	&	I-score: Top Mod.	&	1.00	&	1.00	&	1.00	\\
        \midrule
        Panel B \\
        Training Sample Size: 100	&		&	Test AUC					\\
        Variables:	&		&	Logistic	&	RF	&	NN	\\
        \hline
        100	&	All Var.	&	0.57	&	0.70	&	0.52	\\
        	&	I-score: Top Mod.	&	1.00	&	1.00	&	0.97	\\
        \hline
        200	&	All Var.	&	0.51	&	0.61	&	0.55	\\
        	&	I-score: Top Mod.	&	1.00	&	1.00	&	1.00	\\
        \bottomrule
    \end{tabular}
    \label{tab:simulation-complex-fix-n-change-p-and-fix-p-change-n}
\end{table*}
In this simulation, we suppose the underlying model is not known to us. In this case, we create an artificial example with fixed amount of variables and we let training sample size to vary in 50, 100, or 1000. Since we do not have knowledge of the model, we can either use all variables or we use the proposed method to select the features that have high impact of response variable. This result is presented in Panel A of Table \ref{tab:simulation-complex-fix-n-change-p-and-fix-p-change-n}. We observe similar patterns in previous artificial examples (I - III). In the beginning of Panel A, we have very little observations which pose quite a bit of challenge for common machine learning classifiers to perform well. With the introduction of I-score, we see an immediately improve of the prediction performance. We see further improvement if we raise sample size to 100 and eventually 1,000. We observe all algorithms perform well under large sample size. However, the proposed method I-score does not subject to the harm from the lack of sample size. In Panel B of Table \ref{tab:simulation-complex-fix-n-change-p-and-fix-p-change-n}, we fix sample size in training set to be 100 and we allow number of variables to vary in 100 and 200. We observe the phenomenon that the performance of common machine learning algorithms are quite poor without using I-score. In addition, the performance of all three classifiers produce poorer results when we raise number of variables.

\end{document}